\documentclass[]{typhoon}

\usepackage{amsmath,amssymb}
\usepackage{float}
\usepackage{enumitem}
\usepackage{tabularx}
\usepackage{xspace}
\usetikzlibrary{positioning,arrows.meta,fit,backgrounds,calc,shapes.geometric}

\newcommand{\vtwo}{Typhoon OCR 2\xspace}
\newcommand{\vonefive}{Typhoon OCR 1.5\xspace}
\newcommand{\pipenum}[1]{{\footnotesize\color{black!65}#1}}

\setlist[itemize]{leftmargin=*}
\setlist[enumerate]{leftmargin=*}

\title{Typhoon OCR 2: From Thai Document Transcription to Multilingual, Multi-Task OCR}

\author[1]{Surapon Nonesung}
\author[1]{Natapong Nitarach}
\author[1]{Sittipong Sripaisarnmongkol}
\author[2]{Kunat Pipatanakul}
\author[1]{Warit Sirichotedumrong}
\affiliation[1]{Typhoon, SCB DataX}
\affiliation[2]{AI Singapore}

\abstract{
\vonefive focused on reading Thai documents. However, in real-world workflows, users often need to extract specific information or preserve the structure of an entire page for further processing. \vtwo handles both tasks in a single model, with prompt-driven extraction and structured full-page parsing. Full-page parsing covers Thai, English, and Chinese. We release compact 2B and 0.8B models adapted from open document parsers through supervised fine-tuning on curated Thai and public multilingual data, followed by Group Relative Policy Optimization (GRPO) with task rewards and a rule penalizing unfinished responses. Among evaluated open models, \vtwo 2B scores highest on Thai MDPBench (80.7 versus 68.9 for \vonefive) and on ThaiOCRBench (59.91 versus 40.45), which covers recognition, extraction, and document understanding. The 2B model remains close to its base on English olmOCR-bench (84.7 versus 84.4) and on OmniDocBench, which evaluates English and Chinese together (90.97 versus 91.03). On MDPBench, GRPO reduces the mean share of pages with repetition loops for the 2B model from 3.2\% to 0.7\%, with a smaller and statistically uncertain score gain for the 0.8B model. We release the model weights, Python package, and evaluation suite.
}

\DeclareRobustCommand{\projectresources}{%
  \par\smallskip\noindent
  \begin{tabular}[t]{@{}c@{\hspace{0.4em}}l@{\hspace{1em}}l@{}}
    \raisebox{-1.5pt}{\includegraphics[height=1.05em]{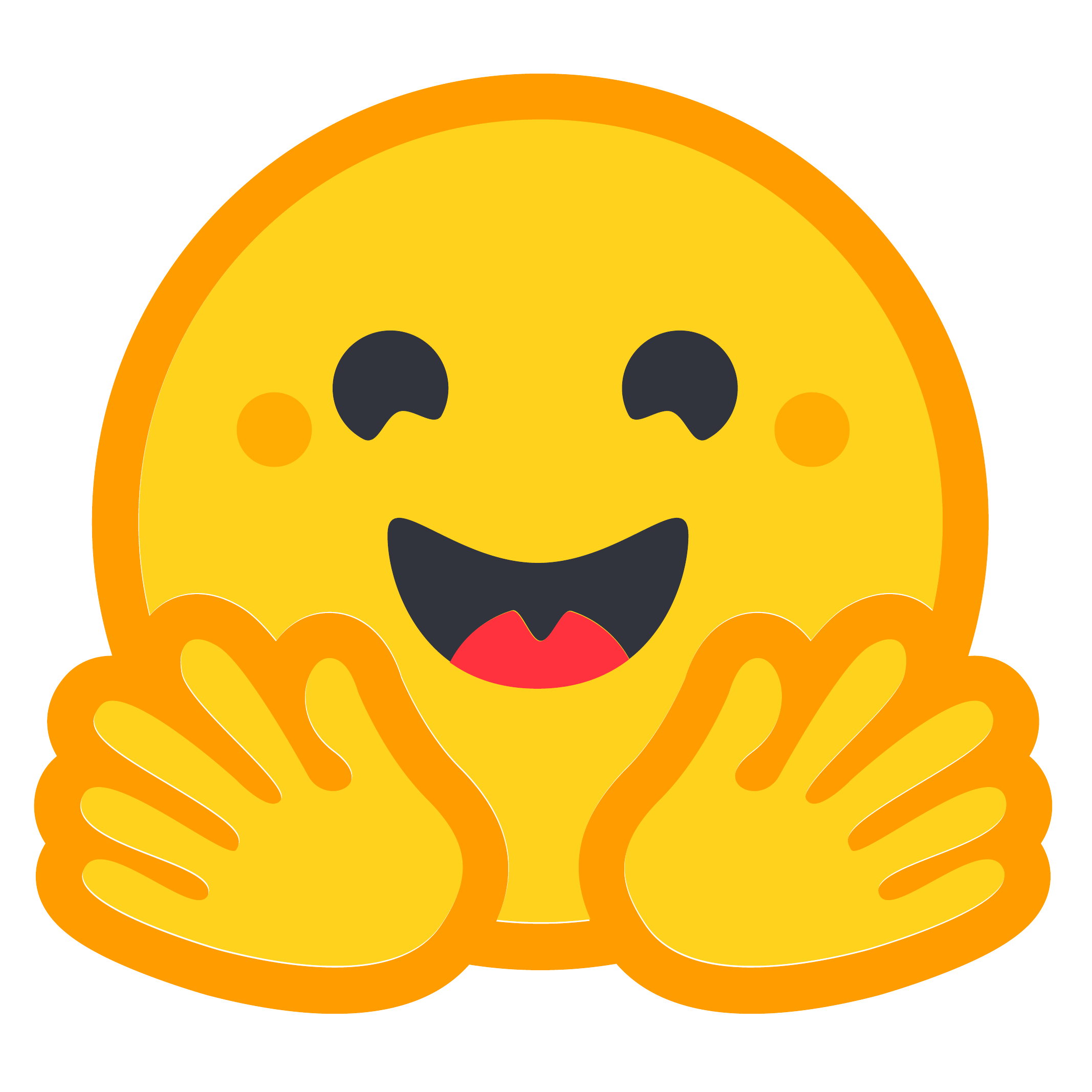}} & Model weights & \href{https://huggingface.co/typhoon-ai/typhoon-ocr2-2b}{2B} \quad \href{https://huggingface.co/typhoon-ai/typhoon-ocr2-0.8b}{0.8B} \\
    \raisebox{-1.5pt}{\includegraphics[height=1.05em]{logos/hf.pdf}} & GGUF & \href{https://huggingface.co/typhoon-ai/typhoon-ocr2-2b-GGUF}{2B} \quad \href{https://huggingface.co/typhoon-ai/typhoon-ocr2-0.8b-GGUF}{0.8B} \\
    \raisebox{-1.5pt}{\includegraphics[height=1.05em]{logos/hf.pdf}} & MLX (8-bit) & \href{https://huggingface.co/typhoon-ai/typhoon-ocr2-2b-MLX-8bit}{2B} \quad \href{https://huggingface.co/typhoon-ai/typhoon-ocr2-0.8b-MLX-8bit}{0.8B} \\
    \raisebox{-1.5pt}{\includegraphics[height=1.05em]{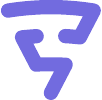}} & Ollama & \href{https://ollama.com/SuraponOCR/typhoon-ocr2-2b}{2B} \quad \href{https://ollama.com/SuraponOCR/typhoon-ocr2-0.8b}{0.8B} \\
    \raisebox{-1.5pt}{\includegraphics[height=1.05em]{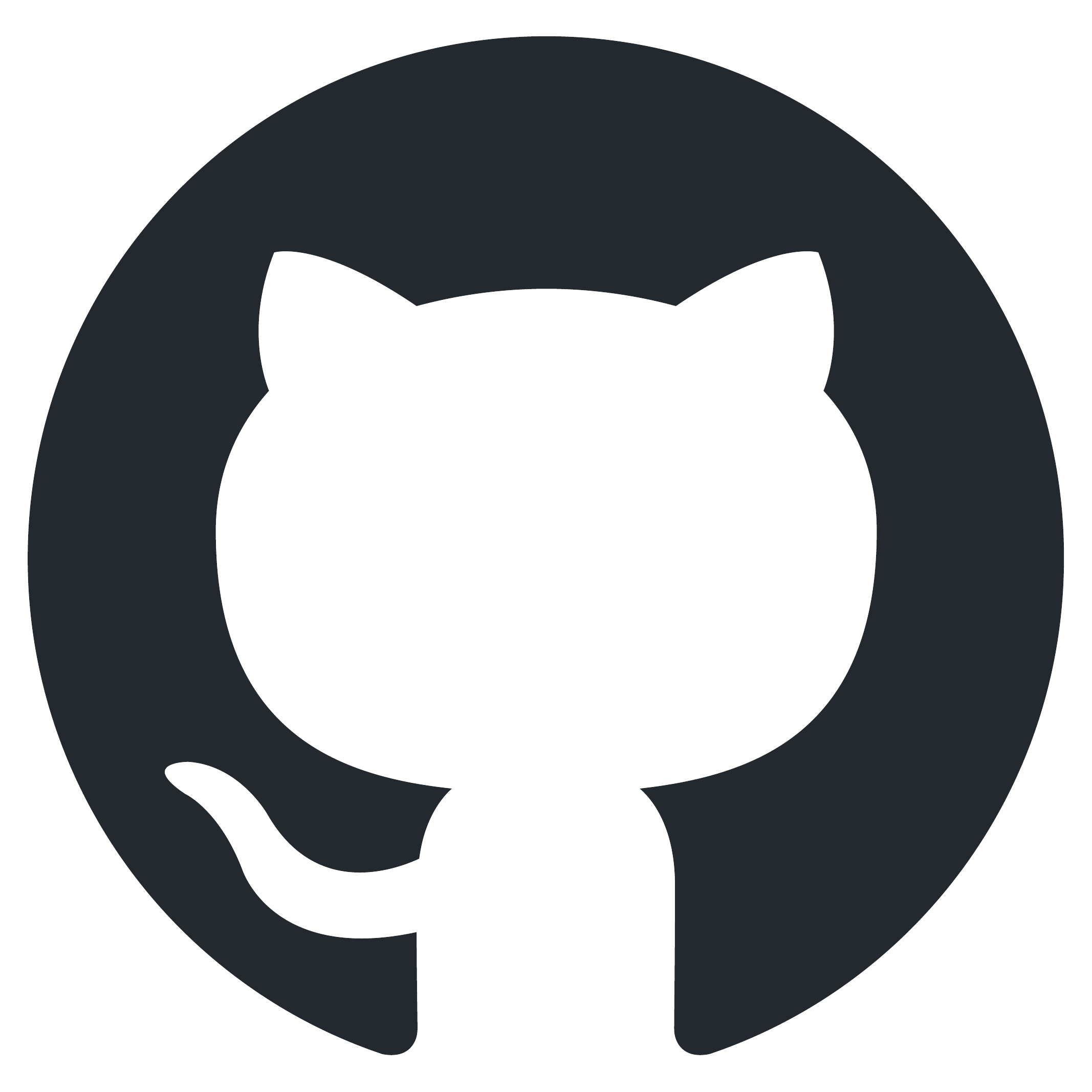}} & Python package & \href{https://github.com/srn-source/typhoon-ocr2}{GitHub} \quad \texttt{pip install typhoon-ocr2} \\
    \raisebox{-1.5pt}{\includegraphics[height=1.05em]{logos/github.pdf}} & Evaluation code & \href{https://github.com/srn-source/evaluation_thai_ocr}{GitHub} \\
    \raisebox{-1.5pt}{\includegraphics[height=1.05em]{logos/typhoon.pdf}} & Demo and API & \href{https://playground.opentyphoon.ai/ocr}{OCR playground} \quad \href{https://docs.opentyphoon.ai}{API documentation} \\
    \raisebox{-1.5pt}{\includegraphics[height=1.05em]{logos/typhoon.pdf}} & Project website & \href{https://opentyphoon.ai}{opentyphoon.ai}
  \end{tabular}%
}
\checkdata[Project Resources]{\projectresources}

\begin{document}
\maketitle
\vspace{-20pt}

\section{Introduction}
\label{sec:intro}
\vonefive \citep{nonesung2026typhoonocropenvisionlanguage} was developed to transcribe Thai documents. A transcript makes the text available, but document workflows often need either a specific piece of information or a structured representation of the whole page. For example, extracting a value from a document can involve running OCR first and then passing the text to a language model to find the value. With \vtwo, the user can ask for that value directly from the page image in a single step.

\begin{figure}[h!]
\centering
\begin{minipage}[c]{0.43\linewidth}
\centering
\includegraphics[width=\linewidth]{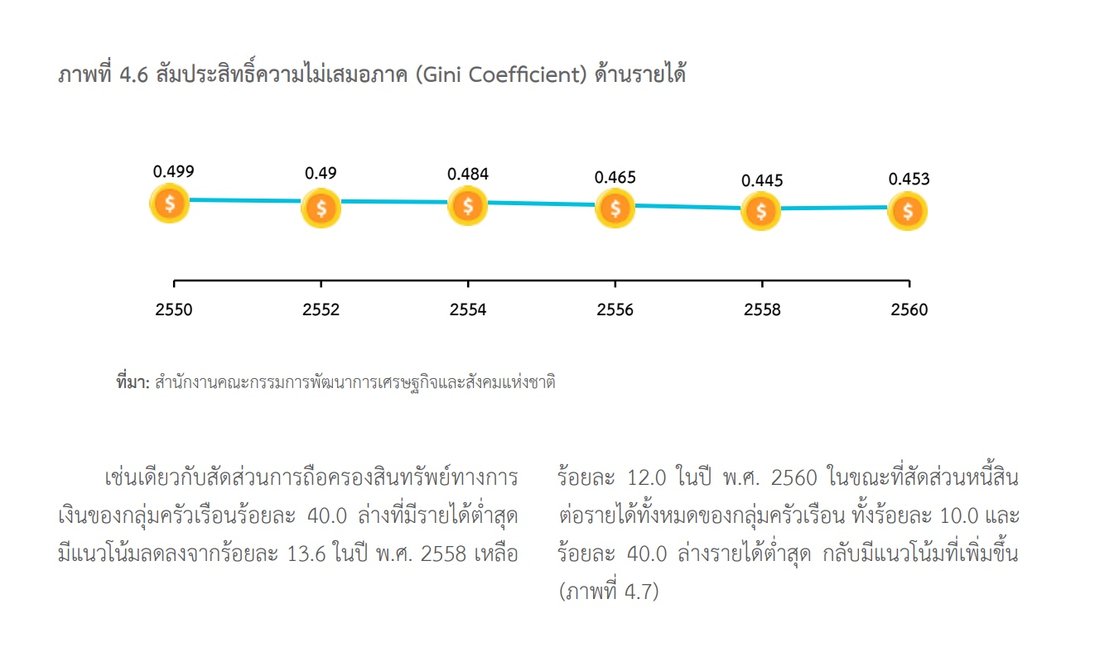}
\end{minipage}\hfill
\begin{minipage}[c]{0.54\linewidth}
\small
\textbf{Requested output.} A dictionary with the chart titles, type, source, and plotted data.\par\smallskip
\textbf{Model response.}\par
\texttt{"chart\_type": "Line chart"}\\
\texttt{"years": ["2550", "2552", "2554",\\
"2556", "2558", "2560"]}\\
\texttt{"Gini Coefficient":\\
{}["0.499", "0.49", "0.484",\\
"0.465", "0.445", "0.453"]}
\end{minipage}
\par\medskip
\includegraphics[height=5.2cm,width=0.47\linewidth,keepaspectratio]{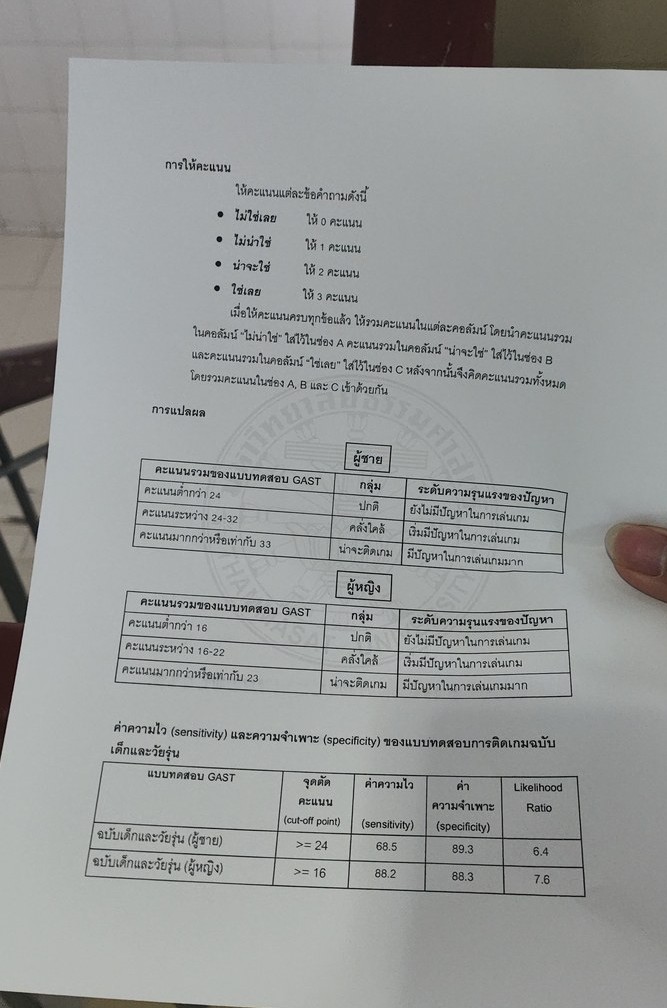}
\hspace{0.04\linewidth}
\includegraphics[height=5.2cm,width=0.47\linewidth,keepaspectratio]{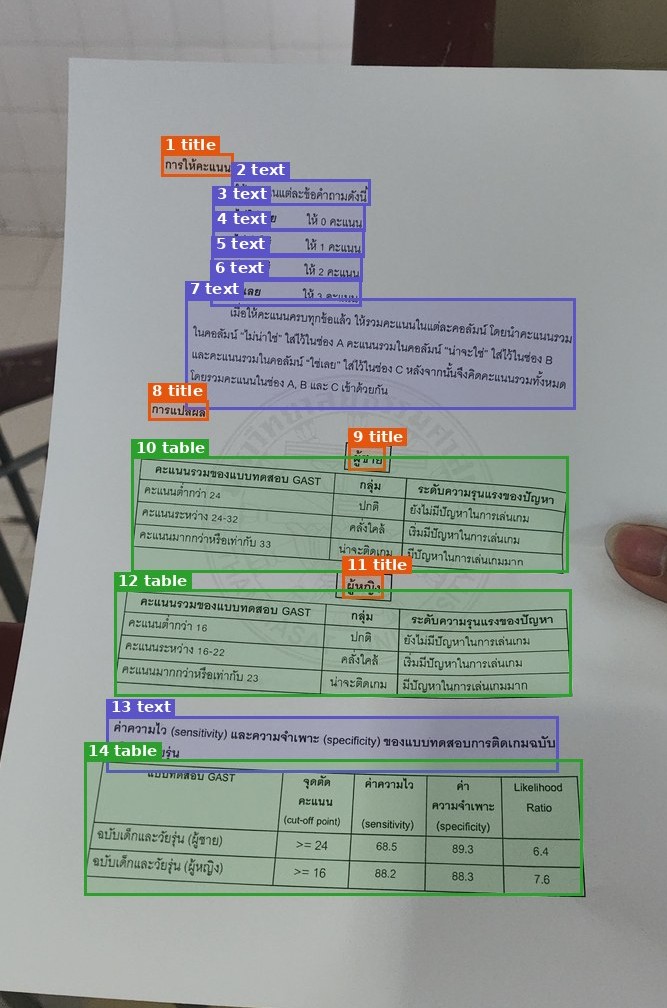}
\caption{\textbf{One model for targeted extraction and full-page parsing.} The top panel shows selected fields returned for a ThaiOCRBench chart prompt. The bottom panel shows an MDPBench page image and the same image with predicted boxes coloured by category and numbered in reading order.}
\label{fig:chart_example}
\end{figure}

We introduce \vtwo, a family of compact open-weight models. At either model size, a single model handles targeted extraction and full-page parsing, with the prompt selecting the task. An extraction prompt asks for specific information, such as values in a chart or fields in a form. For full-page parsing, the model returns ordered layout elements with text, categories, and bounding boxes. \Cref{fig:chart_example} illustrates both outputs on Thai documents.
Thai document understanding remains our focus, and multilingual page parsing supports workflows that also encounter English and Chinese documents. This follows the multilingual approach of other Typhoon work \citep{pipatanakul2024typhoon2familyopen, sirichotedumrong2026typhoonasrrealtimefastconformertransducer}.

We adapt specialized open parsers \citep{huang2026infinityparser2technicalreport,lu2026ovisocr2} with curated Thai data alongside public multilingual data covering English and Chinese, then apply Group Relative Policy Optimization (GRPO) to improve sequence-level performance and reduce failures such as repetition and premature stopping. We evaluate targeted document tasks in Thai and full-page parsing across Thai, English, and Chinese. We also share what we've learned from the development process.

\FloatBarrier
\section{Typhoon OCR 2}
\label{sec:task}
\label{sec:data}

\subsection{Tasks, output formats, and base models}
\label{sec:base}
\label{sec:overview}
The input prompt selects the task for either model. Full-page layout parsing returns a JSON array of elements in reading order. Each element includes a normalized bounding box, a semantic category, and transcribed content. Tables are represented as HTML, formulas as LaTeX, and running text as Markdown. The layout metadata tells an application what each piece of content is, where it appeared on the page, and how it fits into the reading order. An application can use that information to index sections for search, highlight the source of an extracted value, or send individual tables to a data-processing workflow.

Task-specific prompts can request full-page Markdown, boxes without text, text from a specified region, an isolated table or formula, chart data, a document label, an answer to a question about the document, or named fields. The prompt specifies the expected output for the task, such as HTML for a table or a dictionary of chart values. See \Cref{app:prompts} for the page-parsing prompt and output conventions.

We initialize from specialized open document parsers to build on their multilingual document parsing capabilities. The larger model starts from Infinity-Parser2-Flash \citep{huang2026infinityparser2technicalreport}, which uses the same layout JSON format, and the compact model starts from OvisOCR2 \citep{lu2026ovisocr2}. Both are adapted with supervised fine-tuning and GRPO (\Cref{sec:sft,sec:grpo}).

\subsection{Training data}
\label{sec:sft_mix}
\label{sec:public}
We combine rule-filtered \vonefive data with newly labeled in-house Thai examples. Public multitask document data, visual question answering data, and text-only instructions supply additional supervision, with the public sources primarily covering English and Chinese. The mixture includes full-page transcription and layout parsing as well as examples that pair task-specific prompts with targeted outputs. \Cref{tab:sft_sources} summarizes the source mixture before filtering. \Cref{app:data_details} breaks it down by task and document type.

\begin{table}[h!]
\centering
\caption{\textbf{Thai-focused and public multilingual supervision for supervised fine-tuning (SFT).} Filtered \vonefive data and newly labelled Thai-focused examples are combined with public document and instruction data, mainly in English and Chinese. Counts refer to training examples before sequence-length filtering.}
\label{tab:sft_sources}
\small
\begin{tabular}{lrr}
\toprule
\textbf{Source} & \textbf{Examples} & \textbf{Proportion} \\
\midrule
\textit{Thai-focused} & 146,647 & 34.9\% \\
\quad \vonefive data, rule-filtered \citep{nonesung2026typhoonocropenvisionlanguage} & 99,616 & 23.7\% \\
\quad In-house Thai dataset, newly labelled & 47,031 & 11.2\% \\
\addlinespace
\textit{Public, mainly English and Chinese} & 273,680 & 65.1\% \\
\quad Multi-task document data & 237,000 & 56.4\% \\
\quad Visual question answering and text-only instructions & 36,680 & 8.7\% \\
\midrule
\textbf{Total} & \textbf{420,327} & \textbf{100\%} \\
\bottomrule
\end{tabular}
\end{table}

\subsection{Thai label construction}
\label{sec:pipeline}
\label{sec:teacher}
Because public Thai corpora rarely include fine-grained layout and reading-order annotations, we built a multi-stage curation pipeline combining automated labeling, rule and LLM checks, and targeted manual review (\Cref{fig:pipeline}).

\paragraph{Filtering \vonefive data.} The original \vonefive training corpus \citep{nonesung2026typhoonocropenvisionlanguage} contained corrupted files and label noise. We applied deterministic rule checks---verifying syntax, token repetition, LaTeX and table formatting, bounding boxes, and duplicates.

\paragraph{Constructing new Thai annotations.} The new dataset pairs page images with two kinds of supervision: full-page parsing and prompt-driven extraction and visual question answering. For newly collected Thai document pools (PDFs, web images, newspapers, e-books, and \vonefive page samples), we used a multi-stage labeling workflow:
\begin{itemize}[nosep]
    \item \textit{Automated labeling:} Larger vision-language models generate initial annotations: Infinity-Parser2-Pro \citep{huang2026infinityparser2technicalreport} for layout JSON, and other large models for Markdown and key-information extraction.
    \item \textit{Rule checks and LLM verification:} Deterministic rules check candidate annotations, and an LLM judge assigns a quality score by comparing each candidate layout annotation with its page image. Failed samples go back for revision and final correction.
    \item \textit{Targeted bounding-box review:} Because automated checks often miss coordinate inaccuracies, human annotators audit pages with low text-mask coverage and correct inaccurate bounding boxes.
\end{itemize}

\paragraph{Data integrity and contamination check.} Prior to mixing, all sources passed an automated filter for personally identifiable information (PII) using regular expressions, keyword matching, and an LLM judge. We use image hashes to check for exact file overlap with selected benchmarks (\Cref{app:contamination}).

\begin{figure}[h!]
\centering
\resizebox{\linewidth}{!}{%
\begin{tikzpicture}[
  font=\small,
  box/.style={draw=typhoonpurple!75, rounded corners=2pt, fill=white, align=center, text width=4.7cm, inner sep=4pt, minimum height=1.05cm, execute at begin node={\hyphenpenalty=10000\exhyphenpenalty=10000}},
  merge/.style={box, fill=typhoonpurple!10, text width=16.6cm, minimum height=0.85cm},
  outbox/.style={box, fill=typhoonpurple!22},
  note/.style={box, draw=typhoonpurple!55, dashed, font=\footnotesize},
  arr/.style={-{Stealth[length=2.2mm]}, draw=typhoonpurple!80, thick},
  line/.style={draw=typhoonpurple!80, thick},
  elab/.style={font=\footnotesize, text=typhoonpurple!80!black, inner sep=1.5pt},
  lab/.style={font=\small\bfseries, text=typhoonpurple},
  band/.style={draw=typhoonpurple!35, rounded corners=5pt, fill=typhoonpurple!5, inner sep=6pt}
]
\node[box, anchor=north] (a1) at (0,0) {\vonefive training data\\[-1pt]\pipenum{Markdown and VQA; 155,403 examples}};
\node[box, below=0.6cm of a1] (a2) {Deterministic rule checks\\[-1pt]\pipenum{syntax, repetition, LaTeX, tables, boxes, duplicates}};
\node[box, below=0.6cm of a2] (a3) {Examples that pass\\[-1pt]\pipenum{99,616 examples}};
\node[box, anchor=north] (b1) at (5.7,0) {Thai page pools\\[-1pt]\pipenum{PDFs, web images, newspapers, e-books, \vonefive page samples}};
\node[box, below=0.6cm of b1] (b2) {Automatic labelling by larger models\\[-1pt]\pipenum{Infinity-Parser2-Pro (layout JSON); other models (Markdown, KIE)}};
\node[box, below=0.6cm of b2] (b3) {Rule checks and LLM judge\\[-1pt]\pipenum{7,933 layout pages judged; 854 discarded}};
\node[box, below=0.6cm of b3] (b4) {Manual bounding-box review\\[-1pt]\pipenum{low text-mask coverage first; 4,562 pages audited, 1,961 corrected}};
\node[box, below=0.6cm of b4] (b5) {Added: derived tasks, synthetic Thai pages, blank pages\\[-1pt]\pipenum{47,031 examples}};
\node[box, anchor=north] (c1) at (11.4,0) {Multi-task document data\\[-1pt]\pipenum{237,000 examples}};
\node[box, below=0.6cm of c1] (c2) {VQA and text-only instructions\\[-1pt]\pipenum{36,680 examples}};
\coordinate (bus) at ($(c1.east)+(0.4,0)$);
\node[merge, below=0.85cm of b5, xshift=0.4cm] (pii) {PII filter on every source: regular expressions, keyword matching, LLM judge};
\node[outbox, anchor=north] (sft) at ($(pii.south -| b5.south)+(0,-0.75)$) {\textbf{SFT mixture}\\[-1pt]\pipenum{420,327 examples}};
\node[note, anchor=north] (hash) at (sft.north -| a3) {Image-hash check against MDPBench, SEA-DocBench, ThaiOCRBench: no exact match};
\node[outbox, anchor=north] (grpo) at (sft.north -| c1) {\textbf{GRPO prompts}\\[-1pt]\pipenum{12,993: 2,993 Thai pages (annotation score $\ge$ 90), 10,000 public examples unused in SFT}};
\draw[arr] (a1) -- (a2); \draw[arr] (a2) -- (a3);
\draw[arr] (b1) -- (b2); \draw[arr] (b2) -- (b3); \draw[arr] (b3) -- (b4); \draw[arr] (b4) -- (b5);
\draw[arr] (b3.east) -- ++(0.3,0) |- node[elab, pos=0.25, right] {revise} (b2.east);
\draw[line] (c1.east) -- (bus);
\draw[line] (c2.east) -- (c2.east -| bus);
\draw[arr] (a3.south) -- (a3.south |- pii.north);
\draw[arr] (b5) -- (b5.south |- pii.north);
\draw[arr] (bus) -- (bus |- pii.north);
\draw[arr] (pii.south -| sft.north) -- (sft.north);
\draw[arr, dashed] (pii.south -| grpo.north) -- (grpo.north);
\draw[line, dashed, draw=typhoonpurple!60] (hash.east |- sft.west) -- (sft.west);
\node[lab, above=0.15cm of a1.north, anchor=south] (la) {\vonefive data};
\node[lab, above=0.15cm of b1.north, anchor=south] (lb) {In-house Thai dataset};
\node[lab, above=0.15cm of c1.north, anchor=south] (lc) {Public datasets};
\begin{scope}[on background layer]
\node[band, fit=(la)(a1)(a3)(lb)(b1)(b5), label={[lab, font=\normalsize\bfseries]above:Thai-focused sources}] {};
\node[band, fit=(lc)(c1)(c2)(bus), label={[lab, font=\normalsize\bfseries]above:Public sources (English, Chinese)}] {};
\end{scope}
\end{tikzpicture}}
\caption{\textbf{Training data curation for SFT and GRPO.} Legacy Thai data pass deterministic checks. New Thai labels are generated by larger models, verified against the page image, and revised when needed. Annotators prioritize boxes with low text-mask coverage. The supervised mixture combines these sources with public data after privacy screening. Dashed arrows show the separate reinforcement learning (GRPO) prompt set, drawn from accepted Thai pages and public examples held out from SFT.}
\label{fig:pipeline}
\end{figure}

\FloatBarrier
\subsection{Supervised fine-tuning}
\label{sec:sft}

We adapt all parameters of each base parser with next-token cross-entropy on the curated task mixture. This lets the vision encoder and language model jointly adapt to Thai text, layout coordinates, and task-specific output formats. We use a long context window and dynamic image resolution for dense pages with long tables or formulas. Random rotation, downscaling, and photometric adjustments simulate variations in captured documents. Bounding boxes undergo the same geometric transformations as the images. Full training hyperparameters and implementation details are provided in  \Cref{app:training}.

\subsection{GRPO}
\label{sec:grpo}

SFT teaches the model to write a reference answer one token at a time. For document parsing, however, success also depends on the response as a whole: it should cover the page, preserve the requested structure, and finish without repetition loops. Since the SFT objective optimizes next-token predictions along the reference sequence, it does not directly score whether the model's own response completes the task \citep{ranzato2016sequenceleveltrainingrecurrent}. To address this limitation, we apply GRPO \citep{shao2024deepseekmathpushinglimitsmathematical} to post-train the model with sequence-level rewards that directly optimize for task accuracy, output structure, and task completion.

\paragraph{Completion reward.} For a sampled completion $\hat y$ and reference $y$, the task metric $m_{\text{task}}$ measures how well the completion satisfies the request. Each of the 12 tasks in the GRPO prompt set has its own metric (\Cref{tab:reward_tasks}). For full-page layout JSON, it combines category-wise box overlap with text similarity in reading order; tables use TEDS \citep{zhong2020imagebasedtablerecognitiondata}, formulas and Markdown use edit similarity, chart extraction uses value-matching F1, and question answering uses ANLS. Comparing against the full reference also penalizes missing content when a response ends early. The structure term counts triggered rule families, $k(\hat y)$, such as invalid boxes, malformed tables, and heavy repetition (\Cref{tab:reward_rules}). A completion that reaches the token cap without an end-of-sequence token (EOS) receives zero reward, even if it transcribed part of the page correctly:
\begin{equation}
r = \begin{cases}
0 & \text{if $\hat{y}$ reaches the token cap without EOS,}\\
m_{\text{task}}(\hat{y}, y) \cdot \max\!\left(0.6,\; 1 - 0.1\,k(\hat{y})\right) & \text{otherwise,}
\end{cases}
\label{eq:reward}
\end{equation}

Reaching the cap can flag a loop, but a long valid response can also reach it and a repeating response can stop on its own. We therefore detect repetition from the generated text at evaluation time (\Cref{sec:setup}).

\paragraph{Rollouts and optimization.} We sample completions for curated Thai full-page prompts and public prompts held out from SFT. The Thai prompts use layout annotations accepted by the LLM judge during data curation (\Cref{sec:pipeline}) with a quality score of at least 90. These annotations serve as references for scoring the model’s responses. Sampling needs to expose the policy to failures that the reward can penalize. We use a rollout pilot to guide this choice (\Cref{sec:temperature}) and update LoRA adapters \citep{hu2021loralowrankadaptationlarge} with a KL penalty against the SFT policy. \Cref{app:rewards} lists the task metrics and structural rules, and \Cref{app:training} the training settings.

\FloatBarrier
\section{Experiments and results}
\label{sec:results}
\label{sec:eval}

\subsection{Evaluation setup}
\label{sec:setup}
\label{sec:protocol}
\paragraph{Benchmarks and task formats.} We evaluate across five public benchmarks covering full-page parsing, multilingual transcription, and multi-task document understanding (\Cref{tab:benchmarks}). All models decode greedily up to 16,384 new tokens. On the four parsing benchmarks, \vtwo and Infinity-Parser2-Flash take images up to 16.7 MP using the full-page layout JSON prompt, and their JSON is converted to Markdown via Infinity-Parser2's converter. Other baselines run with their author-recommended prompts and post-processing. Our current setup uses vLLM 0.33.0 \citep{kwon2023efficientmemorymanagementlarge} as we observed incorrect outputs from Infinity-Parser2 and \vtwo under vLLM 0.27.0.

\begin{table}[h!]
\centering
\caption{\textbf{Benchmarks for full-page parsing and targeted document tasks.} The page benchmarks cover photographed and digital documents in Thai, English, and Chinese. ThaiOCRBench tests responses to task-specific prompts. Each benchmark uses its official scoring protocol, with arrows marking the preferred direction.}
\label{tab:benchmarks}
\small
\setlength{\tabcolsep}{4pt}
\resizebox{\linewidth}{!}{%
\begin{tabular}{lllll}
\toprule
\textbf{Benchmark} & \textbf{Languages} & \textbf{Scale} & \textbf{Document Type} & \textbf{Primary Metric} \\
\midrule
MDPBench \citep{li2026mdpbenchbenchmarkmultilingualdocument} & EN, TH & 320 pages (160 EN, 160 TH) & Mostly photographed captures & Parsing score (0--100) $\uparrow$ \\
olmOCR-bench \citep{poznanski2025olmocrunlockingtrillionstokens} & EN & 1,403 pages (7,018 tests) & English PDF pages & Test pass rate (\%) $\uparrow$ \\
ThaiOCRBench \citep{nonesung2025thaiocrbenchtaskdiversebenchmarkvisionlanguage} & TH & 2,808 queries (13 tasks) & Multi-task document QA & Mean score over 13 tasks $\uparrow$ \\
SEA-DocBench \citep{yue2026seavisionmultilingualbenchmarkcomprehensive} & EN, TH & 3,091 pages (1,585 EN, 1,506 TH) & Multilingual pages & Edit distance (NED) $\downarrow$ \\
OmniDocBench v1.6 \citep{ouyang2025omnidocbenchbenchmarkingdiversepdf} & EN, ZH & 1,651 pages & Complex PDF layouts & Mean of text, table, and formula scores $\uparrow$ \\
\bottomrule
\end{tabular}%
}
\end{table}

\paragraph{Repetition detection.} Truncation during training is only a proxy for loops (\Cref{sec:grpo}), so on the benchmarks we detect loops from the output text. A page is identified as looping when its output meets any of the following:
\begin{itemize}[nosep]
    \item One line makes up more than half of at least 30 non-empty lines.
    \item A phrase of 6--80 characters repeats 15 or more times in a row.
    \item An output longer than 3,000 bytes compresses with zlib to less than 8\% of its size.
\end{itemize}

\FloatBarrier
\subsection{Thai document understanding and targeted extraction}
\label{sec:main_results}

ThaiOCRBench asks models to respond to task-specific prompts and averages scores across recognition, parsing, and understanding tasks. \vtwo 2B has the highest aggregate score among the evaluated open models, improving from 40.45 for \vonefive to 59.91. Our compact model scores 36.43.

\begin{table}[h!]
\centering
\caption{\textbf{ThaiOCRBench measures broader Thai document understanding.} The score averages recognition, parsing, and targeted document tasks, so \vtwo 2B’s improvement over \vonefive reflects performance across the suite. Scores come from single runs. A dash marks a model without the free-form answering capability needed by the benchmark.}
\label{tab:thai_understanding}
\small
\begin{tabularx}{0.88\linewidth}{@{}>{\raggedright\arraybackslash}Xr@{}}
\toprule
\textbf{Model} & \textbf{ThaiOCRBench average} $\uparrow$ \\
\midrule
\rowcolor{typhoonpurple!20}\multicolumn{2}{l}{\textbf{Open baselines}} \\
Chandra-OCR-2 \citep{paruchuri2026chandra2} & 26.41 \\
dots.mocr \citep{zheng2026multimodalocrparsedocuments} & 37.94 \\
PaddleOCR-VL-1.6 \citep{zhang2026paddleocrvl16expandingfrontierdocument} & -- \\
Infinity-Parser2-Flash \citep{huang2026infinityparser2technicalreport} & 34.98 \\
OvisOCR2 \citep{lu2026ovisocr2} & 10.22 \\
DeepSeek-OCR-2 \citep{wei2026deepseekocr2visualcausal} & 10.10 \\
Nemotron 3.5 Parse \citep{nvidia2026nemotron35parse} & -- \\
OpenThai2.0-OCR-Markdown \citep{openthai2ocr} & -- \\
OpenThai2.1-OCR-Markdown\textsuperscript{\ref{fn:openthai21}} & -- \\
\midrule
\rowcolor{typhoonpurple!20}\multicolumn{2}{l}{\textbf{Typhoon OCR}} \\
\vonefive (2B) \citep{nonesung2026typhoonocropenvisionlanguage} & \underline{40.45} \\
\textbf{\vtwo 0.8B} & 36.43 \\
\textbf{\vtwo 2B} & \textbf{59.91} \\
\bottomrule
\end{tabularx}
\end{table}

\vtwo 2B leads the evaluated open models on ThaiOCRBench with 59.9. The ThaiOCRBench paper reports an average of 77.7 for Gemini 2.5 Pro
\citep{nonesung2025thaiocrbenchtaskdiversebenchmarkvisionlanguage},
17.8 points above our 2B model.

\FloatBarrier
\subsection{Full-page parsing}

\begin{table}[h!]
\centering
\caption{\textbf{Full-page parsing across Thai, English, and Chinese.} \vtwo 2B leads on Thai MDPBench while scoring close to its base on English olmOCR-bench and the combined English--Chinese OmniDocBench. TH, EN, and ZH identify the benchmark languages. SEA-DocBench reports normalized edit distance (NED, lower is better). MDPBench scores for \vtwo and Infinity-Parser2-Flash average four runs. Other scores come from single runs. Bold and underlining mark the best and second-best comparable scores.}
\label{tab:main_results}
\small
\setlength{\tabcolsep}{3pt}
\begin{tabularx}{\linewidth}{@{}>{\raggedright\arraybackslash}Xrrrrrrr@{}}
\toprule
\textbf{Model} & \multicolumn{3}{c}{\textbf{MDPBench} $\uparrow$} & \multicolumn{2}{c}{\shortstack{\textbf{SEA-DocBench}\\\textbf{NED} $\downarrow$}} & \shortstack{\textbf{olmOCR-}\\\textbf{bench} $\uparrow$} & \shortstack{\textbf{OmniDoc}\\\textbf{Bench} $\uparrow$} \\
\cmidrule(lr){2-4}\cmidrule(lr){5-6}
 & TH & EN & All & TH & All & EN & EN+ZH \\
\midrule
\rowcolor{typhoonpurple!20}\multicolumn{8}{l}{\textbf{Open baselines}} \\
Chandra-OCR-2 \citep{paruchuri2026chandra2} & 71.1 & 77.3 & 74.2 & 0.109 & 0.114 & 84.4 & 89.65 \\
dots.mocr \citep{zheng2026multimodalocrparsedocuments} & 73.6 & \underline{78.7} & 76.2 & 0.114 & \underline{0.114} & \textbf{85.2} & 91.09 \\
PaddleOCR-VL-1.6 \citep{zhang2026paddleocrvl16expandingfrontierdocument} & 68.4 & 76.5 & 72.5 & 0.153 & 0.132 & 80.8 & \underline{95.83} \\
Infinity-Parser2-Flash \citep{huang2026infinityparser2technicalreport} & 61.0 & 74.4 & 67.8 & 0.127 & 0.127 & 84.4 & 91.03 \\
OvisOCR2 \citep{lu2026ovisocr2} & 14.7 & \textbf{79.9} & 47.7 & 0.668 & 0.384 & 72.3 & \textbf{96.34} \\
DeepSeek-OCR-2 \citep{wei2026deepseekocr2visualcausal} & 40.7 & 56.9 & 48.9 & 0.144 & 0.136 & 78.2 & 90.05 \\
Nemotron 3.5 Parse \citep{nvidia2026nemotron35parse} & 75.9 & 73.6 & 74.7 & 0.075$^\dagger$ & 0.105$^\dagger$ & 62.7 & 80.13 \\
OpenThai2.0-OCR-Markdown \citep{openthai2ocr} & 69.7 & 67.5 & 68.6 & 0.326 & 0.334 & 51.2 & 62.02 \\
OpenThai2.1-OCR-Markdown\textsuperscript{\ref{fn:openthai21}} & 76.3 & 74.2 & 75.2 & 0.131 & 0.127 & 73.1 & 79.16 \\
\midrule
\rowcolor{typhoonpurple!20}\multicolumn{8}{l}{\textbf{Typhoon OCR}} \\
\vonefive (2B) \citep{nonesung2026typhoonocropenvisionlanguage} & 68.9 & 63.1 & 66.0 & 0.141 & 0.173 & 62.0 & 70.97 \\
\textbf{\vtwo 0.8B} & \underline{76.8} & 76.8 & \underline{76.8} & \textbf{0.100} & \textbf{0.108} & 81.7 & 89.37 \\
\textbf{\vtwo 2B} & \textbf{80.7} & 78.0 & \textbf{79.3} & \underline{0.106} & 0.115 & \underline{84.7} & 90.97 \\
\bottomrule
\end{tabularx}

\smallskip
\begin{minipage}{\linewidth}
\footnotesize $^\dagger$ The scorer omits the formula score for Nemotron 3.5 Parse because the model writes most formulas inline. When all three models are scored on the same components (text, tables, and reading order), Thai NED is 0.075 for Nemotron and 0.074 for both \vtwo 2B and dots.mocr.
\end{minipage}
\end{table}

\paragraph{Thai pages.} On Thai MDPBench, the 2B model improves upon Typhoon OCR 1.5 from 68.9 to 80.7, leading all evaluated open models. The next best scores come from Typhoon OCR 2 0.8B (76.8) and OpenThai2.1-OCR-Markdown (76.3). Our compact model has the lowest Thai SEA-DocBench error among models scored on all four components. Across both MDPBench languages, Typhoon OCR 2 2B scores 79.0 on photographed pages compared to 70.9 for Nemotron 3.5 Parse, while Nemotron leads on digital pages with 86.4 over 81.2 (\Cref{tab:capabilities}).

\paragraph{Multilingual page parsing.} The 2B model improves upon \vonefive on both English MDPBench and olmOCR-bench. On OmniDocBench, its combined score across English and Chinese pages is 90.97, close to its Infinity-Parser2-Flash base at 91.03. OvisOCR2 and PaddleOCR-VL-1.6 remain stronger on complex tables and formulas.

Cross-model comparisons use model-specific prompts and post-processing. The paired intervals in \Cref{tab:paired} describe checkpoint changes, so small gaps in the leaderboard should be interpreted cautiously.

\begin{figure}[h!]
\centering
\resizebox{\linewidth}{!}{%
\begin{tikzpicture}[x=0.18cm, y=0.46cm,
  en/.style={circle, draw=black!55, fill=white, line width=1.1pt, minimum size=5.5pt, inner sep=0pt},
  th/.style={diamond, fill=typhoonpurple, draw=white, line width=0.4pt, minimum size=7.5pt, inner sep=0pt},
  val/.style={font=\scriptsize, text=black!65, inner sep=1pt},
  nm/.style={font=\footnotesize, anchor=east, inner sep=0pt}]
\foreach \v in {40,50,...,80} {
  \draw[black!12] (\v,-0.7) -- (\v,10.6);
  \node[font=\scriptsize, text=black!65, below] at (\v,-0.7) {\v};
}
\draw[black!45] (35,-0.7) -- (88,-0.7);
\node[font=\footnotesize] at (61.5,-2.2) {MDPBench score (higher is better)};
\draw[black!30, dashed] (2,8.5) -- (88,8.5);
\foreach \y/\nm/\en/\th/\enl/\thl in {
  10/{\textbf{\vtwo 2B}}/77.975/80.7/78.0/80.7,
  8/OpenThai2.1-OCR-Markdown/74.2/76.3/74.2/76.3,
  7/Nemotron 3.5 Parse/73.6/75.9/73.6/75.9,
  6/dots.mocr/78.7/73.6/78.7/73.6,
  5/Chandra-OCR-2/77.3/71.1/77.3/71.1,
  4/OpenThai2.0-OCR-Markdown/67.5/69.7/67.5/69.7,
  3/\vonefive/63.1/68.9/63.1/68.9,
  2/PaddleOCR-VL-1.6/76.5/68.4/76.5/68.4,
  1/Infinity-Parser2-Flash/74.4/61.05/74.4/61.0,
  0/DeepSeek-OCR-2/56.9/40.7/56.9/40.7} {
  \node[nm] at (33,\y) {\nm};
  \draw[black!18, line width=2.2pt, line cap=round] (\en,\y) -- (\th,\y);
  \node[en] at (\en,\y) {};
  \node[th] at (\th,\y) {};
  \pgfmathtruncatemacro{\enlow}{\en < \th ? 1 : 0}
  \ifnum\enlow=1
    \node[val, left=3pt] at (\en,\y) {\enl}; \node[val, right=3pt] at (\th,\y) {\thl};
  \else
    \node[val, left=3pt] at (\th,\y) {\thl}; \node[val, right=3pt] at (\en,\y) {\enl};
  \fi
}
\node[nm] at (33,9) {\textbf{\vtwo 0.8B}};
\node[en, minimum size=9.5pt] at (76.75,9) {};
\node[th] at (76.75,9) {};
\node[val, right=5pt] at (76.75,9) {76.8 (EN = TH)};
\node[en] (le) at (37,10) {}; \node[val, right=4pt, font=\footnotesize, text=black] at (le) {English pages};
\node[th] (lt) at (37,9) {}; \node[val, right=4pt, font=\footnotesize, text=black] at (lt) {Thai pages};
\end{tikzpicture}}
\caption{\textbf{Thai scores lag English scores for most general parsers on MDPBench.} Each row compares 160 English pages (circles) with 160 Thai pages (diamonds), mostly photographed documents. \vtwo 2B leads on Thai. The compact model’s markers coincide because its mean scores are equal. Scores for \vtwo and Infinity-Parser2-Flash average four runs. Other scores come from single runs.}
\label{fig:mdp_language_comparison}
\end{figure}

\FloatBarrier
\clearpage
\section{What we learned from training}
\label{sec:analysis}

\subsection{We can improve Thai OCR performance while largely preserving multilingual capabilities}
\label{sec:rq1}

Both model sizes improve substantially on Thai while keeping English MDPBench scores close to their respective base parsers (\Cref{tab:rq1}). For the 2B model, supervised fine-tuning raises Thai MDPBench by 16.8 points in the paired comparison, with no clear change on English (\Cref{tab:paired}). After both training stages, the 2B model remains close to its base on English olmOCR-bench and on OmniDocBench, which evaluates English and Chinese pages together (\Cref{tab:main_results}).

\begin{table}[h!]
\centering
\caption{\textbf{Thai adaptation and page reliability change at different training stages.} Supervised fine-tuning (SFT) improves Thai MDPBench and increases repetition. GRPO lowers the loop count for both model sizes. Scores and looping-page counts refer to the 320-page benchmark. Results average four runs, with loop ranges in parentheses, except for the single OvisOCR2 run.}
\label{tab:rq1}
\small
\setlength{\tabcolsep}{4pt}
\resizebox{\linewidth}{!}{%
\begin{tabular}{lccccccc}
\toprule
\textbf{Checkpoint} & \textbf{Overall}$\uparrow$ & \textbf{EN}$\uparrow$ & \textbf{TH}$\uparrow$ & \textbf{Digital}$\uparrow$ & \textbf{Photo}$\uparrow$ & \textbf{Text edit}$\downarrow$ & \textbf{Looping} \\
\midrule
Infinity-Parser2-Flash (2B base) & 67.8 & 74.4 & 61.0 & 79.2 & 64.5 & 0.227 & 3.0 (2--5) \\
\vtwo 2B, SFT & 77.0 & 75.9 & 78.2 & 81.0 & 76.2 & 0.140 & 10.3 (9--12) \\
\vtwo 2B, GRPO & 79.3 & 78.0 & 80.7 & 81.2 & 79.0 & 0.109 & 2.3 (1--3) \\
\midrule
OvisOCR2 (0.8B base) & 47.7 & 79.9 & 14.7 & 54.3 & 45.5 & 0.469 & 3 \\
\vtwo 0.8B, SFT & 75.2 & 74.8 & 75.7 & 80.3 & 73.6 & 0.149 & 12.8 (12--14) \\
\vtwo 0.8B, GRPO & 76.8 & 76.8 & 76.8 & 82.1 & 75.0 & 0.135 & 8.5 (6--11) \\
\bottomrule
\end{tabular}}
\end{table}

\FloatBarrier
\subsection{GRPO reduces repetition and recovers failed pages}
\label{sec:loops}
\label{sec:decomp}

GRPO brings 2B repetition below the base parser’s level and recovers pages that stopped prematurely. On MDPBench, the average loop count rises from 3.0 to 10.3 after SFT, then falls to 2.3 after GRPO. The lower loop rate on MDPBench suggests that the reliability gains generalize from training prompts to benchmark pages. Across the Thai and English pages in MDPBench, the overall score improves by 2.3 points after GRPO. To understand where the improvement comes from, we group pages by whether they failed before and after GRPO. A page counts as failed if it loops or has normalized text edit distance of at least 0.5.

On average, about 14 pages recover after GRPO. They account for 84\% of the reduction in mean text edit distance (\Cref{tab:decomp}). The recovered pages include repetition loops and incomplete outputs. For example, on one photographed Thai textbook page, the SFT checkpoint returned only the page number, while the GRPO checkpoint transcribed the page (\Cref{fig:repair}). Counting loops alone would miss this failure. The decomposition measures text recovery, while the overall benchmark score also reflects other components.

\FloatBarrier
\subsection{Reward rises and fewer rollouts reach the token cap}
\label{sec:grpo_traj}

\begin{figure*}[h!]
    \centering
    \includegraphics[width=\linewidth,trim=200bp 0 0 0,clip]{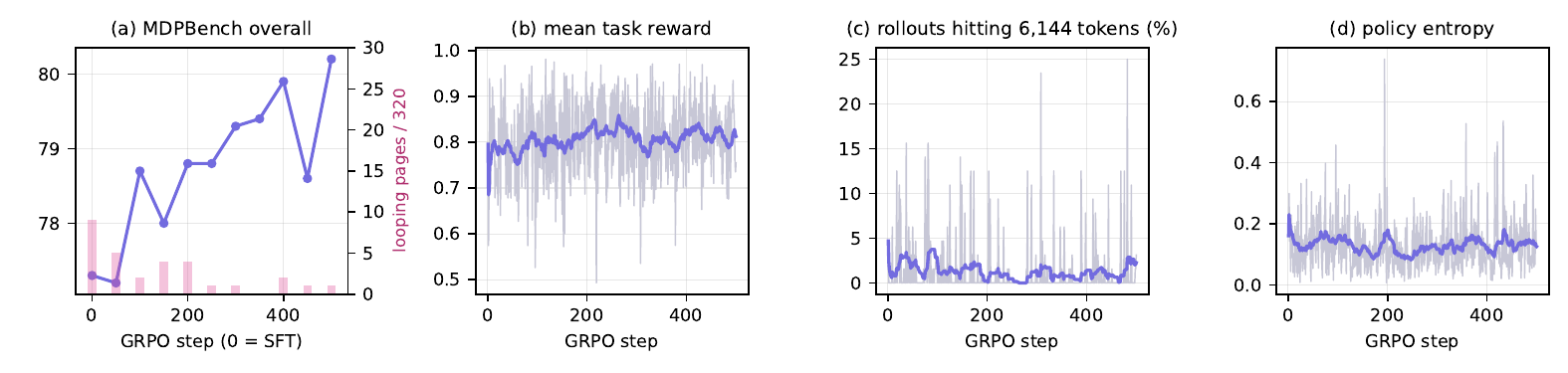}
    \caption{\textbf{Higher rewards and fewer capped completions during 2B GRPO.} Smoothed training logs track mean task reward, the fraction of sampled completions that reach the 6,144-token cap, and policy entropy, a measure of how concentrated the next-token distribution is. Entropy stays broadly stable during the run.}
    \label{fig:dynamics}
\end{figure*}

During 2B GRPO, the model earns higher rewards and produces fewer capped completions while retaining similar sampling diversity. Mean task reward rises from 0.78 to 0.82 while the share of sampled completions reaching the token cap falls from 2.2\% to 0.9\%. Policy entropy remains broadly stable (\Cref{fig:dynamics}). Per task (\Cref{tab:reward_by_task}), the reward rises most for document question answering (0.71 to 0.87) and general questions (0.59 to 0.68), and full-page layout JSON, which makes up 46\% of the rollouts, rises from 0.75 to 0.77. Chart-to-JSON and layout-box rewards fall slightly.
Because each window samples different prompts, these
small changes may reflect sampling variability.

Rollout truncation tracks whether a completion reaches the token cap. Benchmark repetition is measured from the generated content. The rollout-temperature pilot and alternative configurations are reported in \Cref{sec:temperature}.

\FloatBarrier
\section{Limitations and conclusion}
\label{sec:limitations}

\subsection{Limitations}

Structured output is still uneven across tasks. The 2B model scores 21.0 on key-information extraction and 38.1 on document parsing in ThaiOCRBench, and both model sizes trail the strongest baselines on complex tables and formulas. The compact model is better suited to full-page parsing than to targeted document tasks (\Cref{tab:thai_understanding,tab:main_results}). Page parsing also depends on image resolution. Reducing the image limit from 16.7 to 4 megapixels lowered the 2B MDPBench score from 79.8 to 78.2.

Whole-page failures persist after GRPO, especially on photographs. On MDPBench, the released 2B model loops on 1--3 out of 320 pages and the compact model on 6--11 pages(\Cref{tab:rq1,fig:loops_v15}). The checkpoint comparisons measure the combined effect of the GRPO recipe on MDPBench. Because several training settings differ between model sizes, the smaller gain of the compact model cannot be attributed to size alone. The effects of individual reward terms and data sources also remain unmeasured.

Most Thai annotations come from larger models, and manual review concentrated on bounding boxes, leaving room for text and layout errors in the training data. Our overlap check covers byte-identical images in the checked training and benchmark sets (\Cref{app:contamination}).

\subsection{Conclusion}
\label{sec:conclusion}
\vtwo extends Thai-focused OCR to multilingual page parsing and prompt-driven extraction. Among the evaluated open models, \vtwo 2B leads on Thai MDPBench and ThaiOCRBench while retaining scores close to its base on English olmOCR-bench and on the combined English--Chinese OmniDocBench. Leading alternatives remain stronger on complex tables and formulas. Key-information extraction remains difficult on ThaiOCRBench. The compact model offers a smaller option primarily for full-page parsing.

Checkpoint comparisons show that supervised fine-tuning improves Thai parsing and increases whole-page repetition. GRPO reduces looping and recovers failed pages on the 2B model, while the 0.8B score gain is statistically uncertain. The model weights, Python package, and evaluation suite support the use and evaluation of both models.

\section*{Acknowledgments}
This research was supported by SCB DataX, Thailand, and the National Research Foundation, Singapore, under its National Large Language Models Funding Initiative. Any opinions, findings, conclusions, or recommendations expressed in this material are those of the authors and do not reflect the views of the funding organizations.

We thank the authors of Infinity-Parser2 and OvisOCR2, whose open models \vtwo builds on, and the authors of MDPBench, olmOCR-bench, SEA-DocBench, and OmniDocBench, whose public benchmarks made this evaluation possible. Finally, we are deeply grateful to the members of the Typhoon team, past and present. Their ideas, their work and shared commitment to Thai language technology made this work possible, and it belongs to them as much as to its authors.

\bibliographystyle{plainnat}
\bibliography{refs}

\clearpage
\beginappendix
\FloatBarrier
\section{Output format and prompt}
\label{app:prompts}
The doc2json prompt, used for all page-parsing results in this report:
\begin{small}
\begin{verbatim}
- Extract layout information from the provided PDF image.
- For each layout element, output its bbox, category, and the text content
  within the bbox.
- Bbox format: [x1, y1, x2, y2].
- Allowed layout categories: ['header', 'title', 'text', 'figure', 'table',
  'formula', 'figure_caption', 'table_caption', 'formula_caption',
  'figure_footnote', 'table_footnote', 'page_footnote', 'footer'].
- Text extraction and formatting:
  1) For 'figure', the text field must be an empty string.
  2) For 'formula', format text as LaTeX.
  3) For 'table', format text as HTML.
  4) For all other categories (e.g., text, title), format text as Markdown.
- The output text must be exactly the original text from the image, with no
  translation or rewriting.
- Sort all layout elements in human reading order.
- Final output must be a single JSON object.
\end{verbatim}
\end{small}

The raw response places Markdown markers around the JSON list for display. Applications that read JSON directly remove these markers to obtain the data. The example below uses illustrative values.
\begin{small}
\begin{verbatim}
```json
[{"bbox": [62, 48, 938, 96],   "category": "title",  "text": "Annual Report 2025"},
 {"bbox": [62, 120, 938, 410], "category": "text",   "text": "The company ..."},
 {"bbox": [62, 432, 938, 700], "category": "table",  "text": "<table>...</table>"},
 {"bbox": [480, 950, 520, 968], "category": "footer", "text": "12"}]
```
\end{verbatim}
\end{small}

The other tasks of \Cref{sec:task} use short task-specific prompts, for example a one-sentence request for a table in HTML or a formula in LaTeX.

\FloatBarrier
\section{Training data composition}
\label{app:data_details}
\begin{figure*}[h!]
    \centering
    \includegraphics[width=\linewidth]{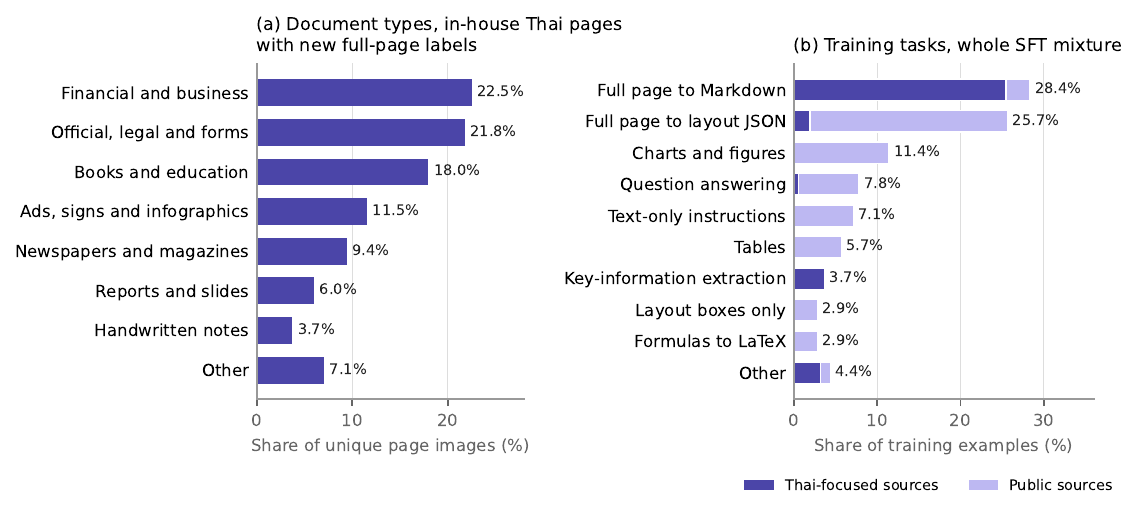}
    \caption{\textbf{Document and task coverage of the supervised training data.} Panel (a) counts unique in-house Thai page images with full-page labels, grouped by document type assigned by the LLM judge. Panel (b) counts training examples by task and source across the full mixture. A page can supply multiple examples. The figure uses separate units to show source-page variety and the supervision the model receives.}
    \label{fig:data_dist}
\end{figure*}

The SFT dataset comprises 420,327 training examples from two Thai-focused and two public sources (\Cref{tab:sft_sources}). \Cref{fig:data_dist} shows the document types of the in-house Thai pages and the task mix of the whole mixture; because several training examples can come from one page, the two panels use different units.

According to the LLM judge (\Cref{sec:pipeline}), the in-house Thai pages with new full-page labels are 49\% scans, 23\% digital documents, 16\% photographs, and 12\% screenshots, and 79\% are in Thai, 11\% in mixed Thai and English, and 10\% in English. Full-page transcription makes up more than half of the mixture. Most Thai-focused examples are Markdown, and Thai-focused sources provide 7.4\% of the layout-JSON examples.

Rule filtering retained 99,616 of 155,403 examples from \vonefive.
The labeling pipeline allows up to two rounds of revision followed by a final correction. Of 7,933 layout pages evaluated by the judge, 3,941 passed directly, 3,138 passed after revision or correction, and 854 were discarded. Reviewers audited 4,562 pages and corrected bounding boxes on 1,961. The in-house mixture contains 47,031 examples after adding derived tasks, synthetic Thai pages, and blank pages.

\FloatBarrier
\section{Training configuration}
\label{app:training}
\begin{table}[h!]
\centering
\caption{\textbf{Full-model supervised adaptation followed by adapter-based GRPO.} The table gives the 2B training settings. The compact model uses the same supervised settings on two GPUs with gradient accumulation adjusted to keep the batch size fixed. Its GRPO configuration is described in \Cref{sec:onebee}.}
\label{tab:hparams}
\small
\begin{tabularx}{\linewidth}{>{\raggedright\arraybackslash}p{3.6cm}>{\raggedright\arraybackslash}X>{\raggedright\arraybackslash}X}
\toprule
 & \textbf{SFT} & \textbf{GRPO} \\
\midrule
Trainable parameters & full model & LoRA, rank 64, $\alpha=128$, on q/k/v/o and MLP projections \\
Optimizer & AdamW, 8-bit states & AdamW \\
Learning rate & $1\times10^{-5}$, one-cycle, 1\% warm-up & $5\times10^{-6}$, cosine \\
Batch & 64 sequences & 8 prompts $\times$ 8 rollouts = 64 completions \\
Sequence / completion length & 14,000 tokens & 6,144 completion tokens \\
Image pixels (processor) & 2,048 to 4,194,304 & 2,048 to 4,194,304 \\
Precision & bf16, FlashAttention-2, gradient checkpointing & bf16, FlashAttention-2, gradient checkpointing \\
Hardware & 4 $\times$ H100, $\sim$81 hours & 2 $\times$ H100 (training) + 2 $\times$ H100 (vLLM), $\sim$5.6 hours \\
\bottomrule
\end{tabularx}
\end{table}

The 2B model initializes from Infinity-Parser2-Flash, which is built on Qwen3.5-2B \citep{qwen3.5}. The 0.8B model initializes from OvisOCR2. We use Axolotl \citep{axolotl} for supervised fine-tuning. For the larger model, GRPO samples at temperature 0.8 and uses a KL coefficient of 0.01.

\paragraph{GRPO prompts and rewards.} The GRPO prompt set, the task metric $m_{\text{task}}$ of each task, and the rules of the structure term $k$ in \Cref{eq:reward} are given in \Cref{app:rewards}.

\paragraph{Optimization.} GRPO uses group-normalized advantages over 8 answers per prompt, the token-level loss of DAPO \citep{yu2025dapoopensourcellmreinforcement} with asymmetric clipping ($\epsilon_{\text{low}}=0.2$, $\epsilon_{\text{high}}=0.28$), and skips batches in which every group has the same reward. Answers come from vLLM on two separate GPUs; the LoRA weights are copied to vLLM periodically, and a token-level importance weight, truncated at 3, corrects for the small lag. The 0.8B GRPO run used temperature 0.6 and one GPU each for sampling and training.

\FloatBarrier
\section{GRPO rewards}
\label{app:rewards}
Every GRPO prompt carries a task name and a reference answer. The reward of a completion is the metric of its task (\Cref{tab:reward_tasks}) multiplied by the structure term of \Cref{eq:reward}, which subtracts 0.1 for each rule family that fires (\Cref{tab:reward_rules}), with a floor of 0.6. A completion that reaches the 6,144-token cap without EOS receives 0, a completion whose output cannot be parsed for its task receives 0, and a completion identical to the reference receives the full metric value of 1 before the structure term. All rewards lie in $[0,1]$. Both model sizes use the same prompts and rewards. The metrics follow Infinity-Parser2 \citep{huang2026infinityparser2technicalreport}, except that formulas use edit similarity instead of CDM, which would need a TeX renderer inside the reward workers.

\begin{table}[h!]
\centering
\caption{\textbf{GRPO scores each answer against a task-specific reference.} Layout parsing combines reading-order text similarity with category-wise box overlap, while other tasks use metrics suited to their outputs. EDS is one minus normalized Levenshtein distance. mIoU$_{\text{cat}}$ averages box-union overlap across categories on a $250\times250$ grid, assigning zero overlap to categories present on only one side. The prompt set combines accepted Thai pages with public examples held out from SFT.}
\label{tab:reward_tasks}
\small
\setlength{\tabcolsep}{4pt}
\begin{tabularx}{\linewidth}{@{}>{\raggedright\arraybackslash}p{3.3cm}r>{\raggedright\arraybackslash}p{2.6cm}>{\raggedright\arraybackslash}X@{}}
\toprule
\textbf{Task (output)} & \textbf{Prompts} & \textbf{Source} & \textbf{Task metric $m_{\text{task}}$} \\
\midrule
Full-page layout JSON & 5,993 & 2,993 in-house Thai, 3,000 public & $0.3\cdot\text{mIoU}_\text{cat} + 0.7\cdot\text{EDS}$ of the element texts in reading order \\
Layout boxes, no text & 500 & public & $\text{mIoU}_\text{cat}$ \\
Full-page Markdown & 500 & public & EDS \\
Table to HTML & 1,000 & public & TEDS on cell content, in either format; $\times\,0.7$ when not in the format the prompt asks for; 0 when the predicted tree is over three times the reference size \\
Table to Markdown & 500 & public & as for Table to HTML \\
Formula to LaTeX & 1,000 & public & EDS on normalized LaTeX (no \texttt{\$}, spacing commands or whitespace) \\
Chart to JSON & 500 & public & F1 over JSON key paths whose values match (numbers within 5\%) \\
Chart to table & 500 & public & F1 over (row, column) cells whose values match (numbers within 5\%) \\
Chart to code & 500 & public & EDS \\
Chemical structure to SMILES & 500 & public & Tanimoto similarity of Morgan fingerprints; 1 for the same canonical SMILES \\
Document question answering & 1,000 & public & ANLS (0 below 0.5) \\
General image questions & 500 & public & ANLS (0 below 0.5) \\
\midrule
Total & 12,993 & & \\
\bottomrule
\end{tabularx}
\end{table}

\begin{table}[h!]
\centering
\caption{\textbf{Structural checks penalize unusable outputs during GRPO.} The same rule families used in supervised data filtering check sampled answers for structural errors. For an answer with positive task reward, each triggered family subtracts 0.1 from the reward multiplier, with a floor of 0.6.}
\label{tab:reward_rules}
\small
\setlength{\tabcolsep}{4pt}
\begin{tabularx}{\linewidth}{@{}>{\raggedright\arraybackslash}p{2.4cm}>{\raggedright\arraybackslash}X>{\raggedright\arraybackslash}p{3.6cm}@{}}
\toprule
\textbf{Rule family} & \textbf{Fires when} & \textbf{Tasks checked} \\
\midrule
Repetition & over 30\% of the 30-character windows repeat, the output (400 characters or more) compresses with zlib to under 6\% of its size, or 10 or more identical lines follow each other & all \\
Table structure & unclosed or wrongly nested HTML table tags, rows of different widths, a row span past the last row, or a ragged Markdown table & layout JSON, layout boxes, tables, chart to table, Markdown \\
LaTeX & empty math span, unbalanced braces or \texttt{\textbackslash left}/\texttt{\textbackslash right}, mismatched \texttt{\textbackslash begin}/\texttt{\textbackslash end}, or a trailing \texttt{\^{}}, \texttt{\_} or backslash & layout JSON, layout boxes, formula, Markdown \\
Bounding boxes & a box outside 0--1000 or with $x_2 < x_1$ or $y_2 < y_1$; pairs of boxes with IoU $\ge 0.9$ (duplicates) or $\ge 0.6$ (heavy overlap) & layout JSON, layout boxes \\
Swapped axes & most text blocks are more than twice as tall as wide ($[y, x]$ order) & layout JSON, layout boxes \\
Categories & a category outside the 13 allowed ones & layout JSON, layout boxes \\
Code syntax & chart code that does not parse as Python & chart to code \\
Thai text & more than two Thai combining marks without a base character, or the replacement character U+FFFD & all except formula \\
\bottomrule
\end{tabularx}
\end{table}

\begin{table}[h!]
\centering
\caption{\textbf{Reward changes vary by task during 2B GRPO.} The table compares mean rewards in the first and last 100 steps. The windows contain different randomly drawn prompts, so small changes on sparsely sampled tasks are difficult to interpret. Rollout counts cover all 500 steps, and table-format tasks are pooled because they share prompt wording. Zero reward includes capped completions, outputs that cannot be parsed, and answers below the ANLS threshold.}
\label{tab:reward_by_task}
\small
\setlength{\tabcolsep}{5pt}
\begin{tabular}{lrccr}
\toprule
 & & \multicolumn{2}{c}{\textbf{Mean reward}} & \\
\cmidrule(lr){3-4}
\textbf{Task} & \textbf{Rollouts} & \textbf{Steps 1--100} & \textbf{Steps 401--500} & \textbf{Zero reward} \\
\midrule
Full-page layout JSON & 14,704 & 0.747 & 0.766 & 3.5\% \\
Layout boxes, no text & 1,264 & 0.907 & 0.872 & 0.6\% \\
Full-page Markdown & 1,200 & 0.950 & 0.957 & 0.2\% \\
Tables (HTML or Markdown) & 3,816 & 0.858 & 0.900 & 0.1\% \\
Formula to LaTeX & 2,496 & 0.872 & 0.849 & 0.4\% \\
Chart to JSON & 1,240 & 0.867 & 0.796 & 0.0\% \\
Chart to table & 1,272 & 0.670 & 0.698 & 22.6\% \\
Chart to code & 1,176 & 0.888 & 0.883 & 0.0\% \\
Chemical structure to SMILES & 1,168 & 0.942 & 0.940 & 1.6\% \\
Document question answering & 2,424 & 0.713 & 0.869 & 18.8\% \\
General image questions & 1,240 & 0.590 & 0.676 & 36.0\% \\
\midrule
All tasks & 32,000 & 0.786 & 0.817 & 5.5\% \\
\bottomrule
\end{tabular}
\end{table}

\FloatBarrier
\section{Contamination checks and reproducibility}
\label{app:contamination}
\paragraph{Contamination check.} We evaluated MD5 image hashes across the primary training directories (approximately 367,000 files) against evaluation images from MDPBench, SEA-DocBench, and ThaiOCRBench, finding zero byte-identical matches. This check detects identical files only and does not identify resized or re-encoded variants. It did not cover visual question-answering images, raw Thai page pools, olmOCR-bench, or OmniDocBench.

\paragraph{Reproducibility.} The complete evaluation suite is publicly available at \url{https://github.com/srn-source/evaluation_thai_ocr}. It runs inference and official benchmark scoring across the models in \Cref{tab:thai_understanding,tab:main_results}. Training uses Axolotl with configurations specified in \Cref{app:training}.

\FloatBarrier
\section{Analysis on parsing accuracy and reliability}
\label{app:stage_evidence}
\label{sec:onebee}

To assess what each training stage changes, we compare checkpoints on the same MDPBench pages with the same runtime and output processing (\Cref{sec:setup}). The paired results show a clear Thai gain after supervised fine-tuning in the 2B model and an overall gain after GRPO. The compact model's overall GRPO gain remains uncertain (\Cref{tab:paired}).

\begin{table}[h!]
\centering
\caption{\textbf{Paired MDPBench comparisons support a Thai SFT gain and a positive overall GRPO gain for the 2B model.} The English SFT change and the compact model’s GRPO gain are uncertain. Differences are in points and use pages scored in every run for both checkpoints (315--316 pages). Parentheses show two-level bootstrap 95\% confidence intervals over pages and runs.}
\label{tab:paired}
\small
\begin{tabular}{lccc}
\toprule
\textbf{Transition} & \textbf{All pages} & \textbf{English} & \textbf{Thai} \\
\midrule
Infinity-Parser2-Flash $\rightarrow$ 2B SFT & $+9.0$ ($+6.1$, $+12.1$) & $+1.5$ ($-1.9$, $+4.9$) & $+16.8$ ($+12.4$, $+21.4$) \\
2B SFT $\rightarrow$ 2B GRPO & $+2.3$ ($+0.9$, $+3.8$) & $+2.1$ ($+0.1$, $+4.2$) & $+2.6$ ($+0.7$, $+4.8$) \\
0.8B SFT $\rightarrow$ 0.8B GRPO & $+1.3$ ($-0.6$, $+3.1$) & $+2.0$ ($-0.4$, $+4.5$) & $+0.6$ ($-2.2$, $+3.4$) \\
\bottomrule
\end{tabular}
\end{table}

The paired score shows whether performance improved on average. To see where it improved, we track which pages recovered, still failed, or failed for the first time after GRPO. Recovery accounts for most of the 2B model's reduction in text error. The compact model also recovers pages, but more new failures offset those gains (\Cref{tab:decomp}). The compact run used a different rollout temperature, hardware allocation, and training schedule, so these comparisons do not isolate the role of model size.

\begin{table}[h!]
\centering
\caption{\textbf{Recovery of failed pages drives most of the 2B text improvement after GRPO.} A page fails if the output loops or its normalized text edit distance is at least 0.5. Share measures each group’s contribution to the net reduction in mean edit distance. It can exceed 100\% when other groups offset the gain. Values average four run pairings over the 316--317 pages scored in both runs of each pairing.}
\label{tab:decomp}
\small
\setlength{\tabcolsep}{4pt}
\resizebox{\linewidth}{!}{%
\begin{tabular}{lcccccc}
\toprule
 & \multicolumn{3}{c}{\textbf{\vtwo 2B}} & \multicolumn{3}{c}{\textbf{\vtwo 0.8B}} \\
\cmidrule(lr){2-4}\cmidrule(lr){5-7}
\textbf{Pages} & \textbf{Pages} & \textbf{Edit, SFT $\rightarrow$ GRPO} & \textbf{Share} & \textbf{Pages} & \textbf{Edit, SFT $\rightarrow$ GRPO} & \textbf{Share} \\
\midrule
Fail under SFT only (repaired) & 13.8 & 0.734 $\rightarrow$ 0.132 & 84\% & 10.5 & 0.730 $\rightarrow$ 0.100 & 155\% \\
\quad of which looped under SFT & 5.5 & 0.614 $\rightarrow$ 0.107 & 29\% & 4.5 & 0.418 $\rightarrow$ 0.099 & 33\% \\
Fail under both & 16.8 & 0.870 $\rightarrow$ 0.810 & 10\% & 22.5 & 0.766 $\rightarrow$ 0.730 & 26\% \\
Fail under GRPO only (new failures) & 1.3 & 0.111 $\rightarrow$ 0.894 & $-10$\% & 6.5 & 0.275 $\rightarrow$ 0.821 & $-92$\% \\
Fail under neither & 285.3 & 0.069 $\rightarrow$ 0.064 & 17\% & 276.8 & 0.073 $\rightarrow$ 0.072 & 11\% \\
\midrule
All pages & 317 & 0.140 $\rightarrow$ 0.109 & 100\% & 316.3 & 0.148 $\rightarrow$ 0.135 & 100\% \\
\bottomrule
\end{tabular}}
\end{table}

\begin{figure}[h!]
    \centering
    \includegraphics[width=\linewidth]{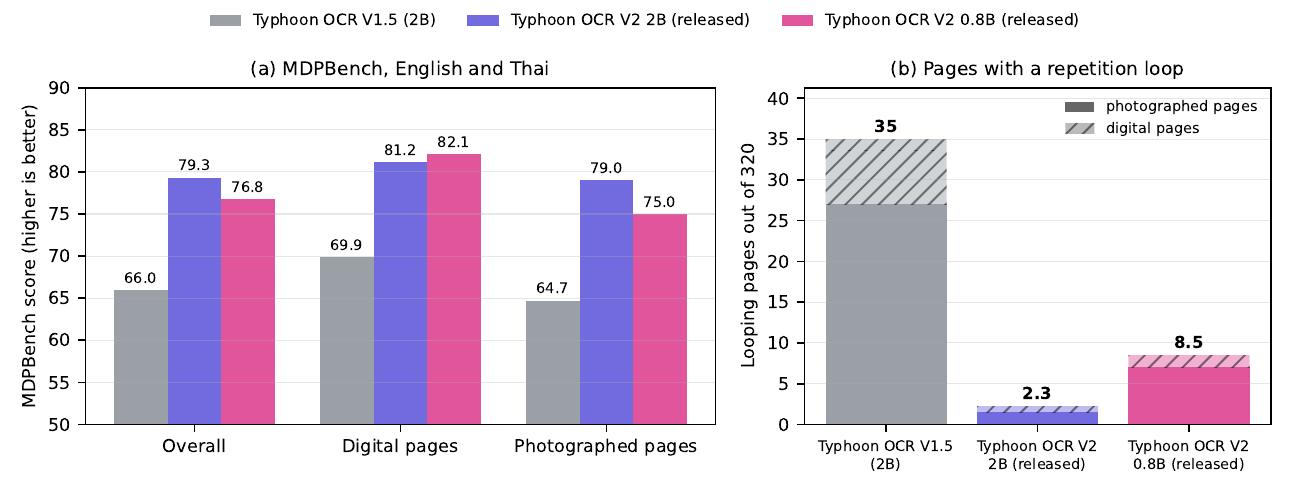}
    \caption{\textbf{Released models improve on \vonefive on photographed pages and loop less often.} Panel (a) shows MDPBench parsing scores overall and by capture type. Panel (b) counts looping outputs for photographed (solid) and digital (hatched) pages. MDPBench contains 320 pages, including 240 photographs. The \vonefive result comes from a single run and the \vtwo results average four runs.}
    \label{fig:loops_v15}
\end{figure}

For a release-level comparison, \Cref{fig:loops_v15} puts the two V2 models alongside \vonefive. The photographed and digital splits show where the gains are largest and where repetition persists.

\FloatBarrier
\section{Pilot experiments on reducing repetition}
\label{sec:temperature}
\label{sec:other}
\label{sec:ablation}
\label{sec:reppen}
Before training, we tested whether rollout sampling would expose the model failures that the reward was designed to penalize. We also tried two ways of addressing repetition: updating all language-model weights during GRPO and discouraging repeated tokens during decoding. These pilots guided the setup.

\paragraph{Choosing a rollout temperature.} GRPO can penalize unfinished responses when they appear in the sampled completions. In the pilot, higher temperatures made capped completions rarer. We chose $T=0.8$ to keep some capped layout responses in the training samples while allowing diverse answers (\Cref{tab:temperature}). In early runs at $T=1.0$, capped completions were rare and greedy repetition did not decrease. Other settings also differed between those runs.

\begin{table}[h!]
\centering
\caption{\textbf{Rollout sampling changes how often GRPO encounters capped responses.} In this pilot, the 2B supervised checkpoint generated four completions per decoding setting for each of 160 prompts. Layout JSON and Markdown responses reached the 6,144-token cap, while short HTML table responses stayed below it. The selected temperature retains some capped layout responses for the reward to penalize.}
\label{tab:temperature}
\small
\setlength{\tabcolsep}{4pt}
\resizebox{\linewidth}{!}{%
\begin{tabular}{lcccc}
\toprule
 & & \multicolumn{3}{c}{\textbf{Task / Output Format}} \\
\cmidrule(lr){3-5}
\textbf{Decoding Strategy} & \textbf{All Tasks} & \textbf{Layout JSON} & \textbf{Markdown} & \textbf{Table HTML} \\
\midrule
Greedy decoding ($T = 0$) & 13.1\% & 15.6\% & 18.8\% & 0.0\% \\
Sampling, $T = 0.6$ & 7.5\% & 12.2\% & 0.8\% & 0.0\% \\
\textbf{Sampling, $T = 0.8$ (Selected)} & \textbf{5.3\%} & \textbf{8.9\%} & \textbf{0.0\%} & \textbf{0.0\%} \\
Sampling, $T = 1.0$ & 1.4\% & 2.3\% & 0.0\% & 0.0\% \\
\bottomrule
\end{tabular}}
\end{table}

\paragraph{Updating all weights during GRPO.} When we updated all language-model weights, the model quickly drifted from the SFT checkpoint and training became unstable. It looped on 35 MDPBench pages, compared with 5 in the LoRA run. This pilot informed our choice to train adapters.

\paragraph{Discouraging repetition at inference time.} We also tried a decoding penalty that makes previously generated tokens less likely. On an earlier SFT checkpoint, a penalty of 1.1 reduced the number of looping pages from 64 to 56. The penalty can also discourage valid repetitions, such as the same digit appearing in several table cells.

\FloatBarrier
\section{Detailed comparison with Nemotron-based parsers}
\label{sec:openthai}
Nemotron 3.5 Parse \citep{nvidia2026nemotron35parse} and its Thai fine-tunes OpenThai2.0-OCR-Markdown \citep{openthai2ocr} and OpenThai2.1-OCR-Markdown\footnote{\label{fn:openthai21}\url{https://huggingface.co/iapp/OpenThai2.1-OCR-Markdown-20261118}, tag \texttt{v2.1.0}, dated 18 November 2026.} follow a different design from ours: a document-to-Markdown pipeline that returns layout boxes, classes, and reading order and does not target free-form question answering. Nemotron 3.5 Parse and OpenThai2.0 keep running headers and footers; OpenThai2.1 leaves them out of the Markdown and writes formulas as LaTeX. \Cref{tab:capabilities} gives the complementary view on the five shared benchmarks of this report, using the recipe from each model card (\Cref{sec:setup}).

\begin{table}[h!]
\centering
\caption{\textbf{Page parsing and output conventions across Typhoon and Nemotron-based models.} Digital and photographed splits expose different parsing strengths. Formula and header scores are also sensitive to formatting choices: OpenThai2.0 writes formulas as plain text, while OpenThai2.1 uses LaTeX and omits running headers and footers. MDPBench scores for \vtwo are four-run means. SEA-DocBench values are for Thai pages, and olmOCR-bench values are single test-file pass rates.}
\label{tab:capabilities}
\small
\setlength{\tabcolsep}{4pt}
\resizebox{\linewidth}{!}{%
\begin{tabular}{lcccc}
\toprule
 & \textbf{\vtwo 2B} & \textbf{Nemotron 3.5 Parse} & \textbf{OpenThai2.0} & \textbf{OpenThai2.1} \\
\midrule
Layout boxes, classes and reading order & yes & yes & yes & yes \\
\addlinespace
\multicolumn{5}{l}{\textit{Page parsing}} \\
\quad MDPBench, Thai pages $\uparrow$ & 80.7 & 75.9 & 69.7 & 76.3 \\
\quad MDPBench, digital pages $\uparrow$ & 81.2 & 86.4 & 77.4 & 81.5 \\
\quad MDPBench, photographed pages $\uparrow$ & 79.0 & 70.9 & 65.7 & 73.1 \\
\quad SEA-DocBench Thai, text NED $\downarrow$ & 0.065 & 0.054 & 0.116 & 0.056 \\
\addlinespace
\multicolumn{5}{l}{\textit{Tables}} \\
\quad OmniDocBench, table TEDS $\uparrow$ & 82.9 & 77.4 & 75.8 & 77.8 \\
\quad SEA-DocBench Thai, table TEDS $\uparrow$ & 0.888 & 0.863 & 0.855 & 0.891 \\
\addlinespace
\multicolumn{5}{l}{\textit{Output conventions (differing across models)}} \\
\quad Formulas as LaTeX: olmOCR-bench, arXiv math $\uparrow$ & 86.4 & 66.4 & 2.3 & 67.1 \\
\quad Formulas as LaTeX: OmniDocBench, formula CDM $\uparrow$ & 94.0 & 71.8 & 24.6 & 66.5 \\
\quad Headers and footers left out (olmOCR-bench file) $\uparrow$ & 94.7 & 20.5 & 23.4 & 77.5 \\
\addlinespace
\multicolumn{5}{l}{\textit{Other tasks}} \\
\quad ThaiOCRBench aggregate $\uparrow$ & 59.9 & not a design target & not a design target & not a design target \\
\bottomrule
\end{tabular}}
\end{table}

\paragraph{Page parsing.} All four models return layout boxes, classes, and reading order. Nemotron 3.5 Parse has the highest score of all evaluated models on digital MDPBench pages (86.4, against 81.2 for \vtwo 2B) and a lower Thai text error on SEA-DocBench (0.054 against 0.065 NED). On photographed MDPBench pages the order reverses: 79.0 for \vtwo 2B, 73.1 for OpenThai2.1-OCR-Markdown, 70.9 for Nemotron 3.5 Parse, and 65.7 for OpenThai2.0-OCR-Markdown. Photographs make up 240 of the 320 MDPBench pages. OpenThai2.1 added phone photos to its training data and gains 7.4 points on them over OpenThai2.0, but remains 5.9 points below \vtwo 2B; on digital pages it scores 81.5, against 81.2 for \vtwo 2B. Its Thai text error on SEA-DocBench (0.056 NED) is close to that of Nemotron 3.5 Parse (0.054).

\paragraph{Output conventions.} Two groups of rows in \Cref{tab:capabilities} depend on output conventions as well as recognition accuracy. OpenThai2.0-OCR-Markdown writes formulas as plain text, while olmOCR-bench, OmniDocBench, and SEA-DocBench score them as LaTeX; on SEA-DocBench its NED is 0.114 over text, tables, and reading order, and 0.334 when the formula component is included. Both Nemotron-based models keep running headers and footers, which one olmOCR-bench test file expects to be removed. OpenThai2.1-OCR-Markdown adopts both conventions, which explains most of its gain over OpenThai2.0 on olmOCR-bench (73.1 against 51.2) and OmniDocBench (79.16 against 62.02); its formula CDM of 66.5 remains well below that of \vtwo 2B (94.0). These rows matter mainly for applications that need formulas in LaTeX or headers removed.

\paragraph{Evaluation scope.} The Nemotron-based models were run once with their published recipes. We did not evaluate \vtwo on the separate Thai test sets reported by OpenThai2.0-OCR-Markdown.

\FloatBarrier
\section{Qualitative examples}
\label{app:examples}
\begin{figure}[h!]
\centering
\includegraphics[height=7.6cm,width=0.48\linewidth,keepaspectratio]{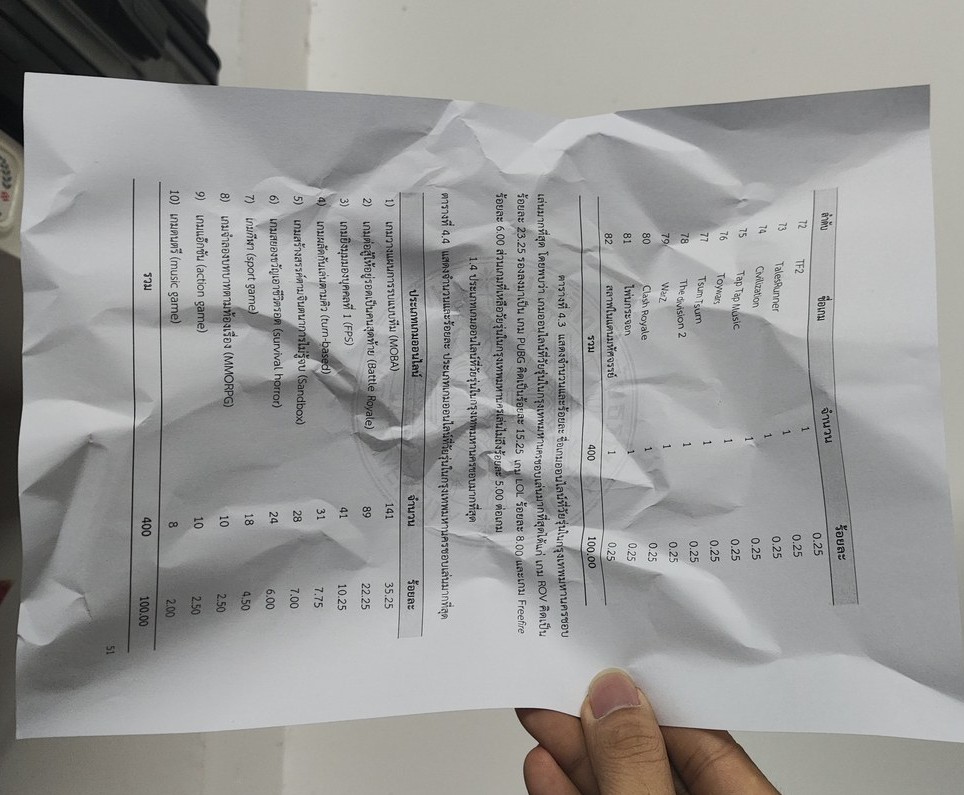}\hspace{0.02\linewidth}%
\includegraphics[height=7.6cm,width=0.48\linewidth,keepaspectratio]{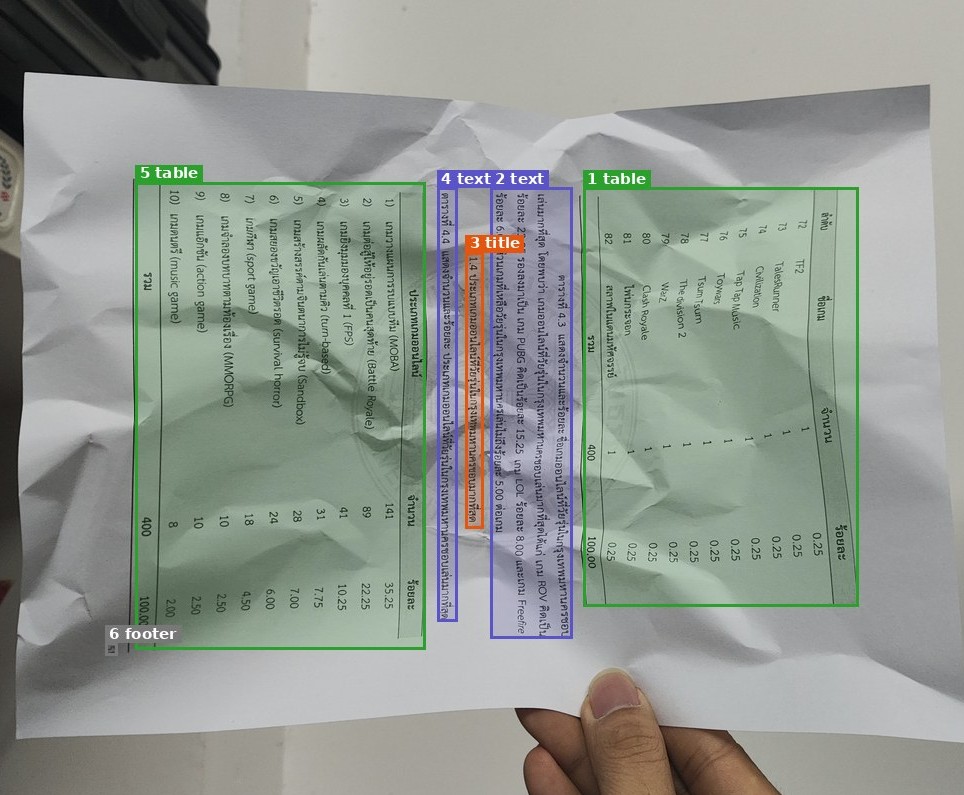}
\begin{tcolorbox}[colback=typhoonpurple!4,colframe=typhoonpurple!40,boxrule=0.4pt,left=4pt,right=4pt,top=2pt,bottom=2pt,fontupper=\footnotesize]
\textbf{Output (complete).} 1.~\texttt{table} [558.8, 374.2, 820.8, 673.2]: \textless{}table\textgreater{}\textless{}tr\textgreater{}\textless{}td\textgreater{}\textthai{ลำดับ}\textless{}/td\textgreater{}\textless{}td\textgreater{}\textthai{ชื่อเกม}\textless{}/td\textgreater{}\textless{}td\textgreater{}\textthai{จำนวน}\textless{}/td\textgreater{}\textless{}td\textgreater{}\textthai{ร้อยละ}\textless{}/td\textgreater{} ... \\ 2.~\texttt{text} [470.8, 374.2, 548.8, 696.2]: \textthai{ตารางที่} 4.3 \textthai{แสดงจำนวนและร้อยละ ชื่อเกมออนไลน์ที่วัยรุ่นในกรุงเทพมหานครชอบ เล่นมากที่สุด โดยพบว่า เกมออนไลน์ที่วัยรุ่นใน} ... \\ 3.~\texttt{title} [446.8, 420.2, 463.8, 617.2]: 1.4 \textthai{ประเภทเกมออนไลน์ที่วัยรุ่นในกรุงเทพมหานครชอบมากที่สุด} \\ 4.~\texttt{text} [419.8, 374.2, 438.8, 684.2]: \textthai{ตารางที่} 4.4 \textthai{แสดงจำนวนและร้อยละ ประเภทเกมออนไลน์ที่วัยรุ่นในกรุงเทพมหานครชอบเล่นมากที่สุด} \\ 5.~\texttt{table} [132.8, 370.2, 408.8, 704.2]: \textless{}table\textgreater{}\textless{}tr\textgreater{}\textless{}td\textgreater{}\textthai{ประเภทเกมออนไลน์}\textless{}/td\textgreater{}\textless{}td\textgreater{}\textthai{จำนวน}\textless{}/td\textgreater{}\textless{}td\textgreater{}\textthai{ร้อยละ}\textless{}/td\textgreater{}\textless{}/tr\textgreater{} ... \\ 6.~\texttt{footer} [103.8, 699.2, 115.8, 708.2]: 51
\end{tcolorbox}
\caption{\textbf{Reading order and table boundaries on a sideways Thai report page.} The box overlay shows the model separating two tables from their introductory text, the title, and the page number despite the 90-degree rotation. Text edit distance is zero across four runs, with table edit distance of 0.025--0.026.}
\label{fig:qual_crumpled}
\end{figure}

\begin{figure}[h!]
\centering
\includegraphics[height=7.6cm,width=0.48\linewidth,keepaspectratio]{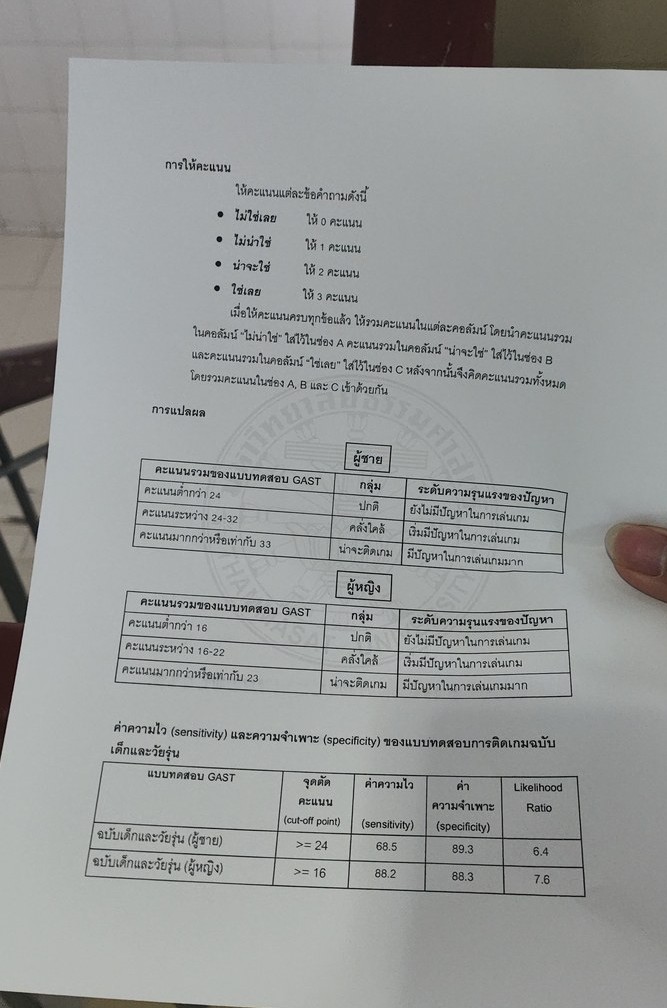}\hspace{0.02\linewidth}%
\includegraphics[height=7.6cm,width=0.48\linewidth,keepaspectratio]{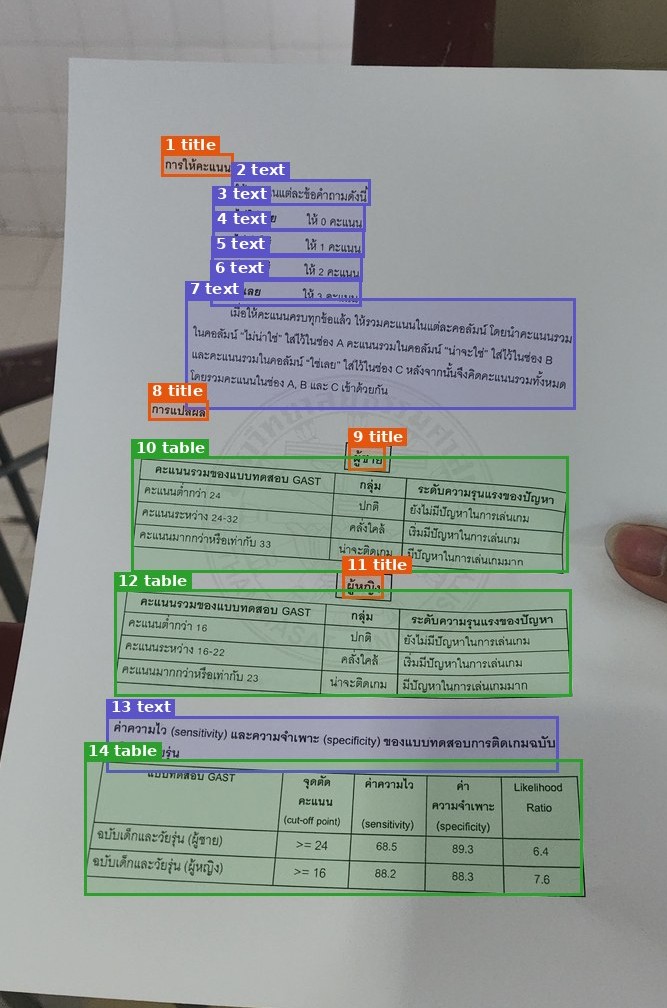}
\begin{tcolorbox}[colback=typhoonpurple!4,colframe=typhoonpurple!40,boxrule=0.4pt,left=4pt,right=4pt,top=2pt,bottom=2pt,fontupper=\footnotesize]
\textbf{Output (first elements).} 1.~\texttt{title} [336.2, 246.3, 405.3, 262.2]: \textthai{การให้คะแนน} \\ 2.~\texttt{text} [403.2, 264.4, 535.2, 283.2]: \textthai{ให้คะแนนแต่ละข้อคำถามดังนี้} \\ 3.~\texttt{text} [385.2, 281.6, 530.3, 301.8]: • \textthai{ไม่ใช่เลย ให้} 0 \textthai{คะแนน} \\ 4.~\texttt{text} [385.2, 299.6, 529.2, 320.2]: • \textthai{ไม่น่าใช้ ให้} 1 \textthai{คะแนน} \\ 5.~\texttt{text} [384.2, 317.4, 527.2, 337.6]: • \textthai{น่าจะใช้ ให้} 2 \textthai{คะแนน} \\ 6.~\texttt{text} [383.2, 334.6, 526.2, 355.2]: • \textthai{ใช้เลย ให้} 3 \textthai{คะแนน} \\ 7.~\texttt{text} [359.2, 349.6, 729.8, 429.2]: \textthai{เมื่อให้คะแนนครบทุกข้อแล้ว ให้รวมคะแนนในแต่ละคอลัมน์ โดยนำคะแนนรวมในคอลัมน์} “\textthai{ไม่น่าใช้}” \textthai{ใส่ไว้ในช่อง} ... \\ 8.~\texttt{title} [323.8, 422.2, 380.8, 436.6]: \textthai{การแปลผล} \\ 9.~\texttt{title} [514.7, 454.8, 549.2, 472.2]: \textthai{ผู้ชาย} \\ 10.~\texttt{table} [308.2, 462.8, 723.2, 545.2]: \textless{}table\textgreater{}\textless{}tr\textgreater{}\textless{}td\textgreater{}\textthai{คะแนนรวมของแบบทดสอบ} GAST\textless{}/td\textgreater{}\textless{}td\textgreater{}\textthai{กลุ่ม}\textless{}/td\textgreater{}\textless{}td\textgreater{}\textthai{ระดับควา} ... \\ \ldots\ (14 elements in total; elements 11, 12, 13, 14 omitted)
\end{tcolorbox}
\caption{\textbf{Layout and reading order on a handheld Thai report page.} The overlay shows separate regions for list items, explanatory text, three tables, and the short headings above the first two tables. The text matches the reference across four runs, while table edit distance ranges from 0.055 to 0.056.}
\label{fig:qual_handheld}
\end{figure}

\begin{figure}[h!]
\centering
\includegraphics[height=7.6cm,width=0.48\linewidth,keepaspectratio]{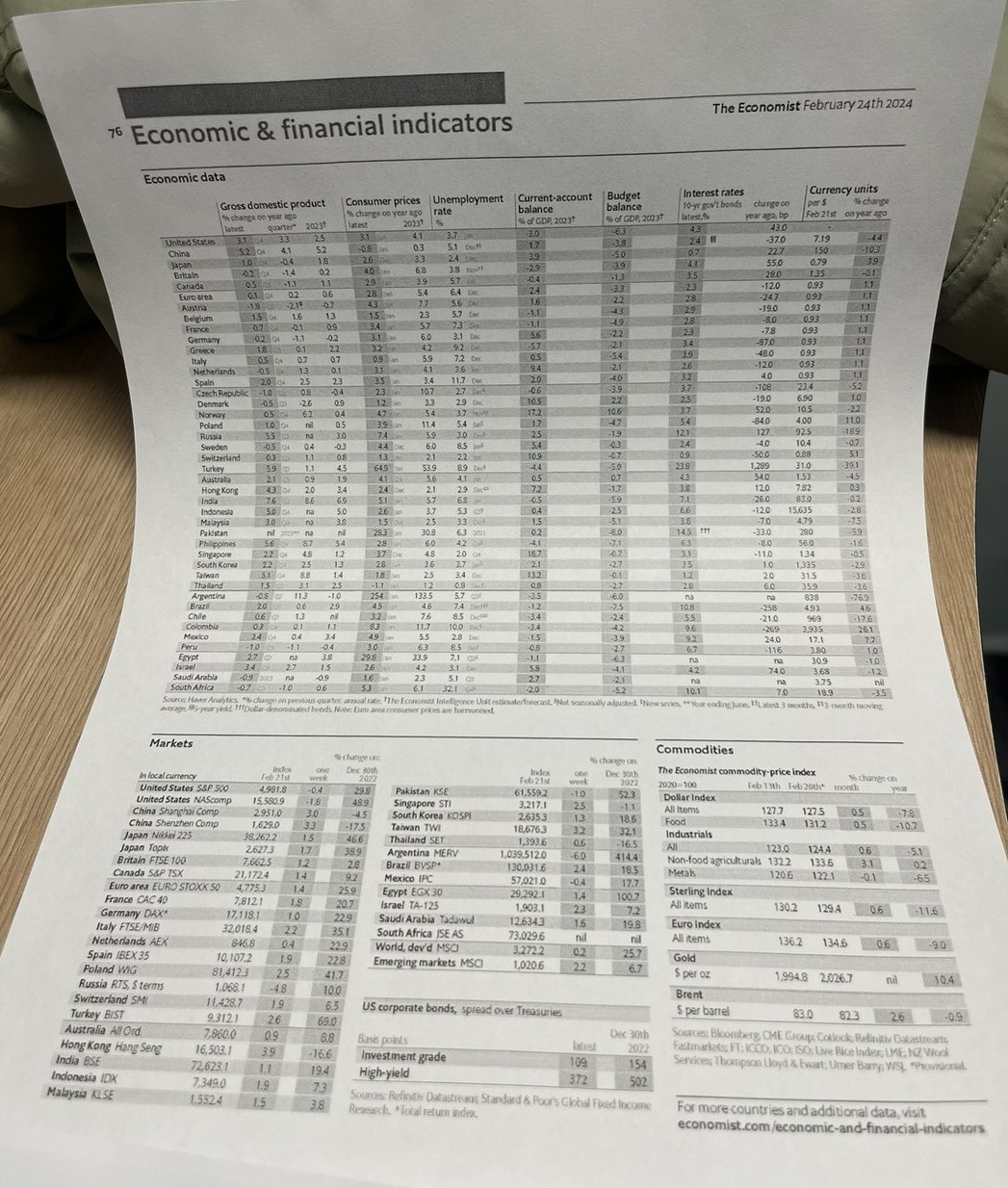}\hspace{0.02\linewidth}%
\includegraphics[height=7.6cm,width=0.48\linewidth,keepaspectratio]{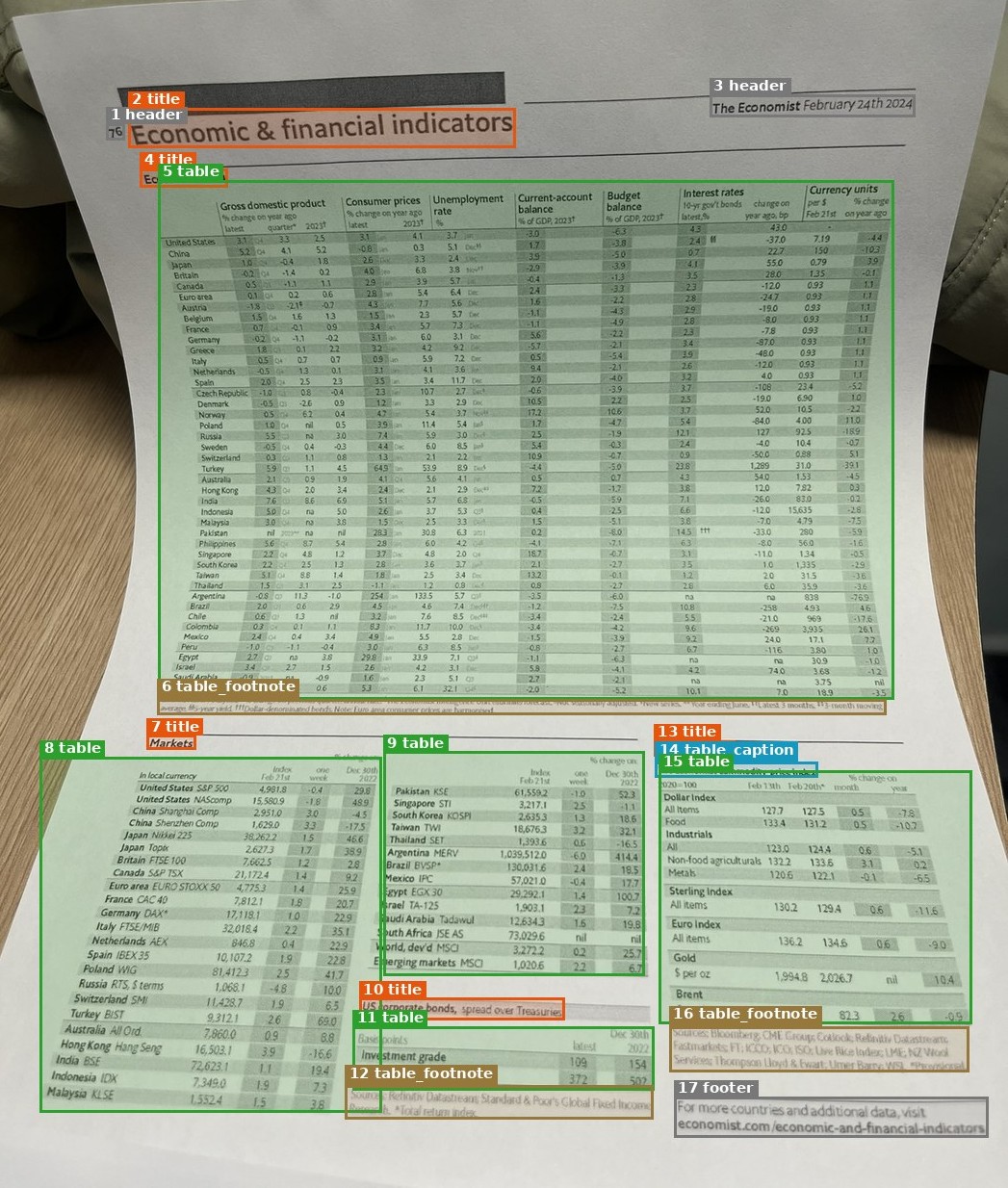}
\begin{tcolorbox}[colback=typhoonpurple!4,colframe=typhoonpurple!40,boxrule=0.4pt,left=4pt,right=4pt,top=2pt,bottom=2pt,fontupper=\footnotesize]
\textbf{Output (selected elements).} 1.~\texttt{header} [108.2, 210.2, 126.2, 222.2]: 76 \\ 2.~\texttt{title} [129.7, 198.7, 512.7, 228.2]: Economic \& financial indicators \\ 3.~\texttt{header} [705.2, 188.7, 908.2, 205.2]: The Economist February 24th 2024 \\ 4.~\texttt{title} [141.7, 243.7, 228.2, 257.2]: Economic data \\ 5.~\texttt{table} [159.7, 252.2, 887.2, 637.2]: \textless{}table\textgreater{}\textless{}tr\textgreater{}\textless{}td\textgreater{}\textless{}/td\textgreater{}\textless{}td colspan="2"\textgreater{}Gross domestic product \% change on ... \\ 6.~\texttt{table\_footnote} [158.7, 637.2, 879.2, 649.2]: Source: Haver Analytics. *\% change on previous quarter, annual rate. †The Economist Intell ... \\ 7.~\texttt{title} [148.2, 664.2, 196.2, 674.7]: Markets \\ 8.~\texttt{table} [42.7, 680.2, 380.2, 944.2]: \textless{}table\textgreater{}\textless{}tr\textgreater{}\textless{}td\textgreater{}\textless{}/td\textgreater{}\textless{}td\textgreater{}\textless{}/td\textgreater{}\textless{}td colspan="3"\textgreater{}\% change on:\textless{}/td\textgreater{}\textless{}/tr\textgreater{}\textless{}tr ... \\ 13.~\texttt{title} [649.7, 668.2, 730.2, 679.7]: Commodities \\ 14.~\texttt{table\_caption} [651.2, 684.2, 812.2, 695.7]: The Economist commodity-price index \\ 17.~\texttt{footer} [669.7, 932.2, 980.2, 962.2]: For more countries and additional data, visit economist.com/economic-and-financial-indicat ... \\ \ldots\ (17 elements in total; elements 9, 10, 11, 12, 15, 16 omitted)
\end{tcolorbox}
\caption{\textbf{Recognizing text and separating dense tables on a bent English magazine page.} The overlay shows five distinct tables and the surrounding captions, source notes, header, and footer. Text edit distance is zero across four runs, while table edit distance of 0.128--0.148 shows remaining errors in table parsing.}
\label{fig:qual_magazine}
\end{figure}

\begin{figure}[h!]
\centering
\begin{minipage}[c]{0.40\linewidth}
\centering
\includegraphics[width=\linewidth,height=7.5cm,keepaspectratio]{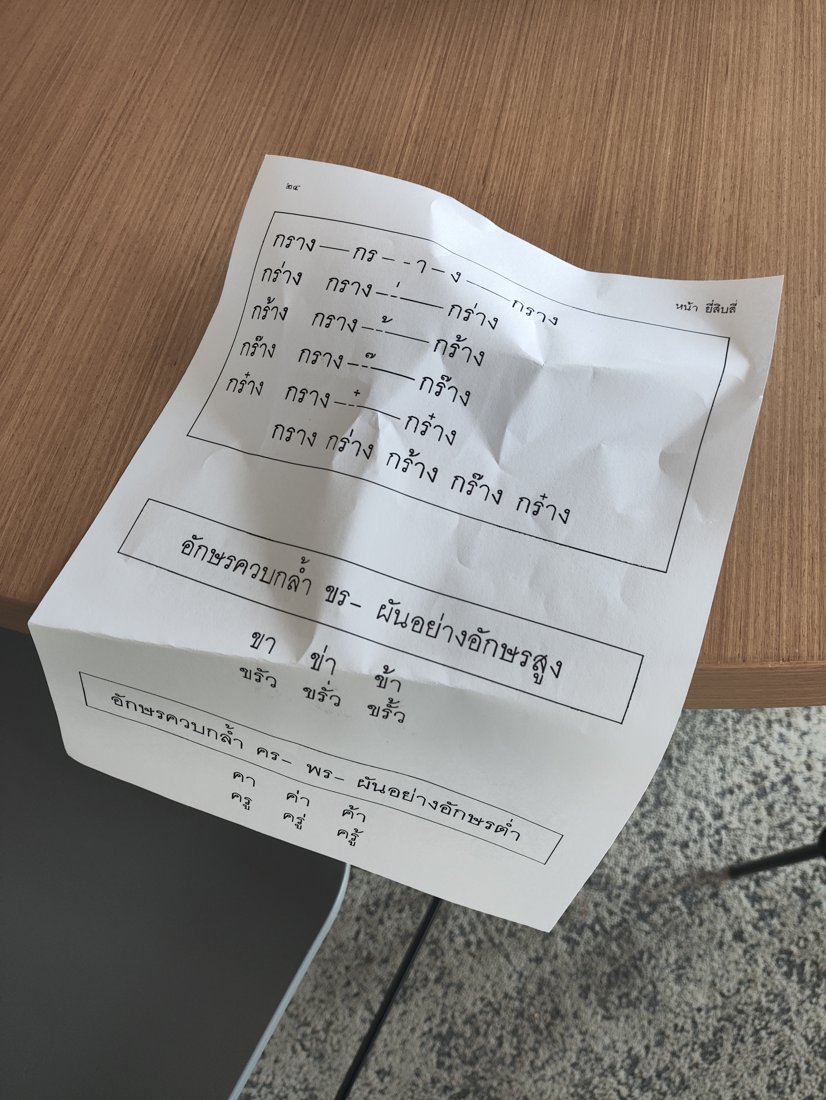}
\end{minipage}\hfill
\begin{minipage}[c]{0.56\linewidth}
\begin{tcolorbox}[colback=typhoonpurple!4,colframe=typhoonpurple!40,boxrule=0.4pt,left=4pt,right=4pt,top=2pt,bottom=2pt,fontupper=\footnotesize]
\textbf{SFT checkpoint, complete output.} \textthai{๒๔ หน้า ยี่สิบสี่}\\[2pt]
\textbf{GRPO model, first lines.}\\
\textthai{กราง --- กร --- ำ --- ง --- กราง}\\
\textthai{กร่าง กราง --- ๑ --- กร่าง}\\
\textthai{กร้าง กราง --- ๒ --- กร้าง}\\
\textthai{กร้าง กราง --- ๓ --- กร้าง}\\
\textthai{กร้าง กราง --- ๔ --- กร้าง}\\
\textthai{กราง กร่าง กร้าง กร้าง กร้าง}\\
\textthai{อักษรควบกล้ำ} \ldots
\end{tcolorbox}
\end{minipage}
\caption{\textbf{GRPO recovers a page that the supervised model ends prematurely.} On this photographed Thai textbook page, the supervised checkpoint returns only the page number. The GRPO checkpoint transcribes the page, reducing normalized text edit distance from 1.00 to 0.03 in all four runs. This example illustrates a recovered failure that a repetition detector would miss.}
\label{fig:repair}
\end{figure}

\vtwo is meant for documents as people capture them, not only for born-digital pages. \Cref{fig:qual_crumpled,fig:qual_handheld,fig:qual_magazine} show full-page parsing by \vtwo 2B on MDPBench phone photographs with hard capture conditions: crumpled, bent, and hand-held paper, a sideways photo, and a slight tilt. Each figure shows the input photo (left) and the predicted layout (right): one box per element, coloured by category and numbered in reading order, followed by the first elements of the raw output. We checked the boxes and categories of these pages by eye. The boxes are axis-aligned rectangles, so on strongly tilted photos they fit rotated text loosely and can overlap; the examples were chosen among photos where this does not happen. Outputs come from vLLM 0.33.0 in its default mode with the settings of \Cref{sec:setup}; each is identical to one of the four MDPBench runs, and the captions give the page's MDPBench scores over the four runs. \Cref{fig:repair} illustrates a page that failed under SFT and was repaired under GRPO (\Cref{sec:decomp}).

\end{document}